\documentclass[journal]{IEEEtran}

\usepackage{epsfig}
\usepackage{graphicx}
\usepackage{amsmath}
\usepackage{amssymb}
\usepackage{graphics} 
\usepackage{mathptmx} 
\usepackage{booktabs}
\usepackage{multirow}
\usepackage{color}
\usepackage{mathrsfs}
\usepackage{float}
\usepackage{subfigure}
\usepackage{url}

\ifCLASSOPTIONcompsoc
  \usepackage[nocompress]{cite}
\else
  \usepackage{cite}
\fi

\ifCLASSINFOpdf
\else
\fi
\begin{document}
%
\title{Deep Direct Visual Odometry}
%
%
%

\author{Chaoqiang~Zhao,
        Yang~Tang,~\IEEEmembership{Senior Member,~IEEE,}
        Qiyu~Sun,
        and Athanasios V. Vasilakos

\thanks{This work was supported in part by National Natural Science Foundation of China (Basic Science Center Program: 61988101), International (Regional) Cooperation and Exchange Project(6 1720106008), the Program of Shanghai Academic Research Leader under Grant No. 20XD1401300, in part by the Programme of Introducing Talents of Discipline to Universities (the 111 Project) under Grant B17017 (\textit{Corresponding author: Yang Tang}).}
\thanks{C. Zhao, Y. Tang, and Q. Sun are with the Key Laboratory of Smart Manufacturing in Energy Chemical Process, Ministry of Education, East China University of Science and Technology, Shanghai, 200237, China (e-mail: zhaocqilc@gmail.com (C. Zhao); yangtang@ecust.edu.cn, tangtany@gmail.com (Y. Tang).}
\thanks{A. V. Vasilakos is with the School of Electrical and Data Engineering, University of Technology Sydney, Australia, with the College of Mathematics and Computer Science, Fuzhou University, Fuzhou, 350116, China, and with the Department of Computer Science, Electrical and Space Engineering, Lulea University of Technology, Lulea,97187, Sweden (Email:th.vasilakos@gmail.com)}
}

%
%

\markboth{}%
{Shell \MakeLowercase{\textit{et al.}}: Bare Demo of IEEEtran.cls for IEEE Journals}
%



\maketitle

\begin{abstract}
Traditional monocular direct visual odometry (DVO) is one of the most famous methods to estimate the ego-motion of robots and map environments from images simultaneously. However, DVO heavily relies on high-quality images and accurate initial pose estimation during tracking. 
With the outstanding performance of deep learning, previous works have shown that deep neural networks can effectively learn 6-DoF (Degree of Freedom) poses between frames from monocular image sequences in the unsupervised manner. However, these unsupervised deep learning-based frameworks cannot accurately generate the full trajectory of a long monocular video because of the scale-inconsistency between each pose. To address this problem, we use several geometric constraints to improve the scale-consistency of the pose network, including improving the previous loss function and proposing a novel scale-to-trajectory constraint for unsupervised training. We call the pose network trained by the proposed novel constraint as TrajNet.
In addition, a new DVO architecture, called deep direct sparse odometry (DDSO), is proposed to overcome the drawbacks of the previous direct sparse odometry (DSO) framework by embedding TrajNet. Extensive experiments on the KITTI dataset show that the proposed constraints can effectively improve the scale-consistency of TrajNet when compared with previous unsupervised monocular methods, and integration with TrajNet makes the initialization and tracking of DSO more robust and accurate.
\end{abstract}


%
\IEEEpeerreviewmaketitle

\section{INTRODUCTION}



Simultaneous localization and mapping (SLAM) and visual odometry (VO) \cite{forster2016svo,sappa2008efficient} play an important role in various fields, such as virtual/augmented reality \cite{lategahn2014vision} and autonomous driving \cite{liu2017robust}. Because of its real-time performance and low computational complexity, VO has attracted increasing attention in robotic pose estimation and environmental mapping \cite{scaramuzza2011visual}.
In recent years, different kinds of approaches have been proposed to solve VO problems, including direct methods \cite{engel2014lsd,engel2017direct} and indirect methods \cite{forster2016svo,mur2017orb}. In contrast to indirect methods, direct methods address the pose by using photometry information directly and eliminate the need to calculate feature descriptors and perform feature matching.
As a result, direct methods are more robust in the case of motion blur or highly repetitive texture \cite{engel2017direct}. Nevertheless, compared to indirect methods, direct methods suffer from the following two major drawbacks \cite{von2020gn}: 1) direct methods are sensitive to the photometric changes between frames, and 2) direct methods need a good initialization.

\begin{figure}[t]
	\centering
	\subfigure[DDSO on Seq. 10]{
		\includegraphics[width = \columnwidth]{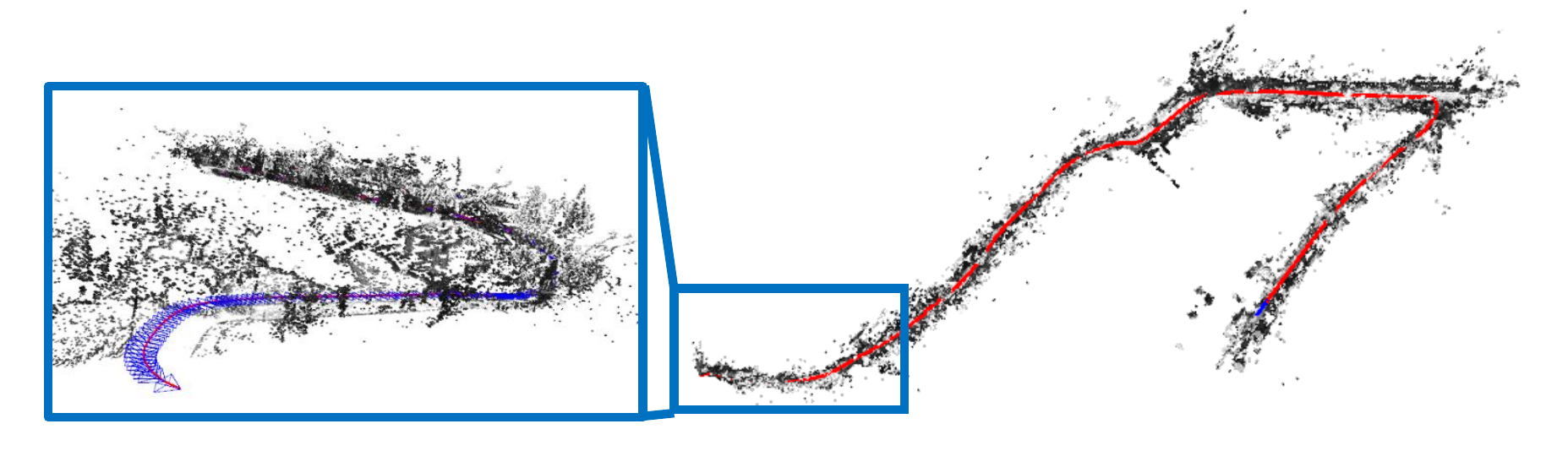}
	}
	\subfigure[DSO \cite{engel2017direct} on Seq. 10]{
		\includegraphics[width = \columnwidth]{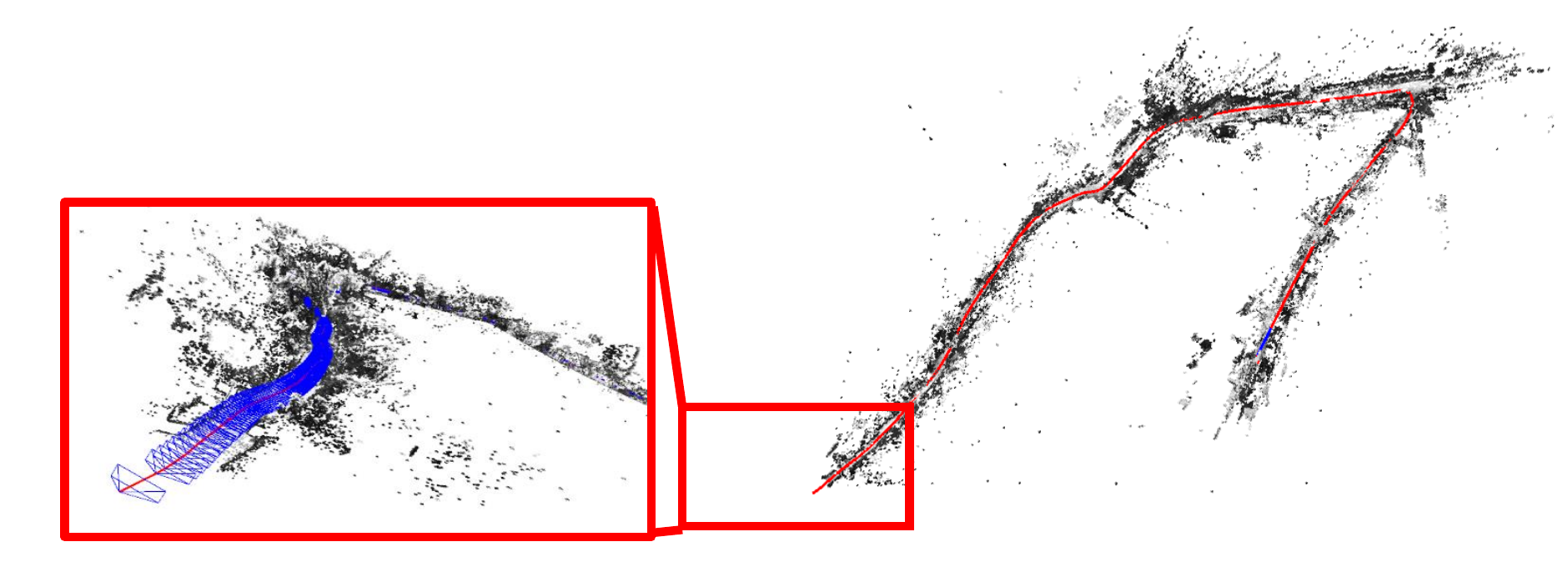}
	}
	\caption{Trajectories of DDSO and DSO \cite{engel2017direct} running on KITTI odometry sequence 10. The proposed DDSO is more robust than DSO \cite{engel2017direct} in initialization. (a). DDSO successfully completes the initialization, even if there are large motions and visual field changes (winding road) in the initial stage. (b). DSO \cite{engel2017direct} can only finish the initialization process only when there is a small view change (straight road).}
	\label{fig:fig1}
\end{figure}

For example, as one of the most famous direct methods, direct sparse odometry (DSO) \cite{engel2017direct} is sensitive to the photometric changes between frames, and robust initialization and accurate tracking are difficult in challenging environments, as shown in Fig. \ref{fig:fig1}.
During tracking, the accuracy of poses predicted by DSO \cite{engel2017direct} depends on the image alignment algorithm. This algorithm obtains the inter-frame pose by optimizing the initial pose provided by the constant motion model. The constant motion model assumes that the current inter-frame pose is the same as the last inter-frame pose, so this initial pose improved by the constant motion model is not accurate enough. In addition, in the initialization process, the constant motion model is not applicable due to the lack of prior motion information, and the pose is initialized as a unit matrix. It is not accurate to use the unit matrix as the initial pose of the inter-frame, thereby leading to the failure of the initialization process. Hence, in this paper, we use deep learning-based pose estimation to provide more accurate initial poses and improve the initialization and tracking of DSO \cite{engel2017direct}. The poses estimated by deep neural networks are used to replace the unit matrix in the initialization process and replace the constant motion model in the tracking process.

Compared with traditional VO methods, deep learning-based VO models learn the geometry from massive images directly, estimate the depth in an end-to-end manner, and do not rely on high-precision correspondence of features or high-quality images \cite{kendall2015posenet}. Recently, deep learning-based models for VO have been proposed and trained via geometric constraints in an unsupervised manner \cite{zhou2017unsupervised}. Because unsupervised methods \cite{zhou2017unsupervised,bian2019depth} do not rely on expensive ground truth but monocular videos, they have recently received much attention.
In this unsupervised framework, the pose network is designed to be jointly trained with a depth estimation network. The outputs of these two networks construct the corresponding relationship of pixels between sequential frames based on projection. Then, the photometric error between corresponding pixels is regarded as the main supervised signal for training.
However, due to the lack of scale information, these unsupervised methods \cite{zhou2017unsupervised,yin2018geonet} trained on monocular videos suffer from monocular ambiguity and per-pose scale inconsistency, which makes it difficult to accurately generate the full trajectory of a monocular video. To obtain the scale information, Zhan et al. \cite{zhan2018unsupervised} train their pose network with stereo image sequences in an unsupervised manner, and the scale information can be learned from stereo image pairs. To address the per-pose scale inconsistency problem, Bian et al. \cite{bian2019depth} and Zhao et al. \cite{zhao2020masked} propose a novel geometric constraint to align the scale information between different depth maps estimated by depth network. Since the pose and depth networks are tightly coupled during training, the scale consistency of the pose network is improved by this constraint at the same time.

In this paper, based on the previous network architecture and unsupervised framework \cite{zhou2017unsupervised,yin2018geonet,bian2019depth,zhao2020masked}, we address the per-pose scale inconsistency problem by proposing a novel scale-consistent constraint, called the pose-to-trajectory constraint. We also improve the view reconstruction constraint in \cite{zhou2017unsupervised,yin2018geonet,bian2019depth,zhao2020masked} to better adapt to the scale consistency of TrajNet. The proposed constraint as well as the improvement of the view reconstruction constraint align the scale between each pose to improve the scale-consistency of TrajNet. Therefore, the TrajNet trained by the proposed constraint can accurately estimate the trajectory over the long monocular video.

In addition, we incorporate TrajNet with DSO \cite{engel2017direct}, called deep direct sparse odometry (DDSO).
First, the poses estimated by TrajNet are optimized by local optimization in DSO \cite{engel2017direct}, and the 3D geometry of scenes can be visualized with the mapping thread of DSO \cite{engel2017direct}. In addition, DSO \cite{engel2017direct} is capable of obtaining a more robust initialization and accurate tracking with the aid of TrajNet, as shown in Fig. \ref{fig:fig1}.

The main contributions of this work are as follows:

\begin{itemize}
	
	\item To improve the scale-consistency of TrajNet, this paper improves the previous view reconstruction constraint and proposes a novel pose-to-trajectory constraint to constrain the training of TrajNet. Compared with previous works \cite{zhou2017unsupervised,yin2018geonet,bian2019depth,zhao2020masked}, TrajNet can predict a more accurate trajectory for a long monocular video.
	\item To address the drawbacks of DSO, a novel direct VO framework incorporating DSO \cite{engel2017direct} with TrajNet, called DDSO, is proposed to improve the robustness and accuracy of DSO \cite{engel2017direct}. DDSO achieves more robust initialization and accurate tracking than DSO \cite{engel2017direct}.
	\item Extensive experiments on the KITTI dataset \cite{geiger2013vision} show the effectiveness of our proposed methods. Moreover, this paper demonstrates the feasibility of learning-based pose estimation for improving traditional DVO framework.
	
\end{itemize}

 This work is organized as follows: In section II, the related works on monocular VOs are discussed. Section III introduces the unsupervised TrajNet and DDSO frameworks in detail. Section IV shows the experimental results of TrajNet and DDSO on the KITTI dataset. Finally, this study is concluded in section V.

\section{RELATED WORK}

\subsection{Geometry-based framework}

 SLAM and VO can be divided into filter-based \cite{ullah2020simultaneous} and optimization-based \cite{engel2017direct} methods based on the back-end and divided into feature-based \cite{mur2017orb}, semi-direct \cite{forster2016svo} and direct \cite{engel2014lsd} methods based on the visual frond-end.
 The traditional sparse feature-based method \cite{mur2017orb} is used to estimate the pose from a set of keypoints by minimizing the reprojection error. Because of the heavy cost of feature extraction and matching, this method has a low speed and poor robustness in low-texture scenes. Engel et al. \cite{engel2014lsd} present a semi-dense direct framework that employs photometric errors as a geometric constraint to estimate the motion. However, this method optimizes the structure and motion in real-time and tracks all pixels with gradients in the frame, which is computationally expensive. Therefore, a direct and sparse method is then proposed in \cite{engel2017direct}, which has been manifested as more accurate than \cite{engel2014lsd}, by optimizing the poses, camera intrinsics and geometry parameters based on a nonlinear optimization framework. Then, an approach with a higher speed that combines the advantage of feature-based and direct methods is designed by Forster et al. \cite{forster2016svo}. However, these approaches in \cite{engel2017direct,forster2016svo} are sensitive to photometric changes and rely heavily on accurate initial pose estimation, which makes initialization difficult and prone to fail in the case of large motion or photometric changes. In this paper, to improve the robustness and accuracy of the initialization and tracking in DSO \cite{engel2017direct}, a novel framework, called DDSO, is designed by leveraging TrajNet into the previous DSO framework\cite{engel2017direct}.

\subsection{Deep learning-based architecture}

With the development of deep neural networks, end-to-end pose and depth estimation has achieved great progress  \cite{zhao2020monocular,zhang2020when}. 
In recent years, unsupervised-based approaches have become more popular because of freeing from the ground truth. Zhou et al. \cite{zhou2017unsupervised} design an unsupervised framework consisting of a pose network and a depth network. Based on the output of the two networks, the view reconstruction loss is calculated and regarded as the main supervisory signal. Therefore, their pose and depth networks can be trained by monocular videos without any ground truth labels. Yin et al. \cite{yin2018geonet} propose the jointly unsupervised learning framework, including monocular depth, optical flow and camera pose, and their work demonstrates the advantages of exploiting geometric relationships over different isolated tasks. More recently, many methods \cite{tosi2019learning, poggi2020uncertainty} have been proposed to improve the accuracy of monocular depth estimation by infusing traditional stereo knowledge or estimating the uncertainty of depth estimation.

\subsection{Scale-inconsistency in monocular unsupervised framework}
\begin{figure}[t]
	\centering
	\includegraphics[width = 0.9\columnwidth]{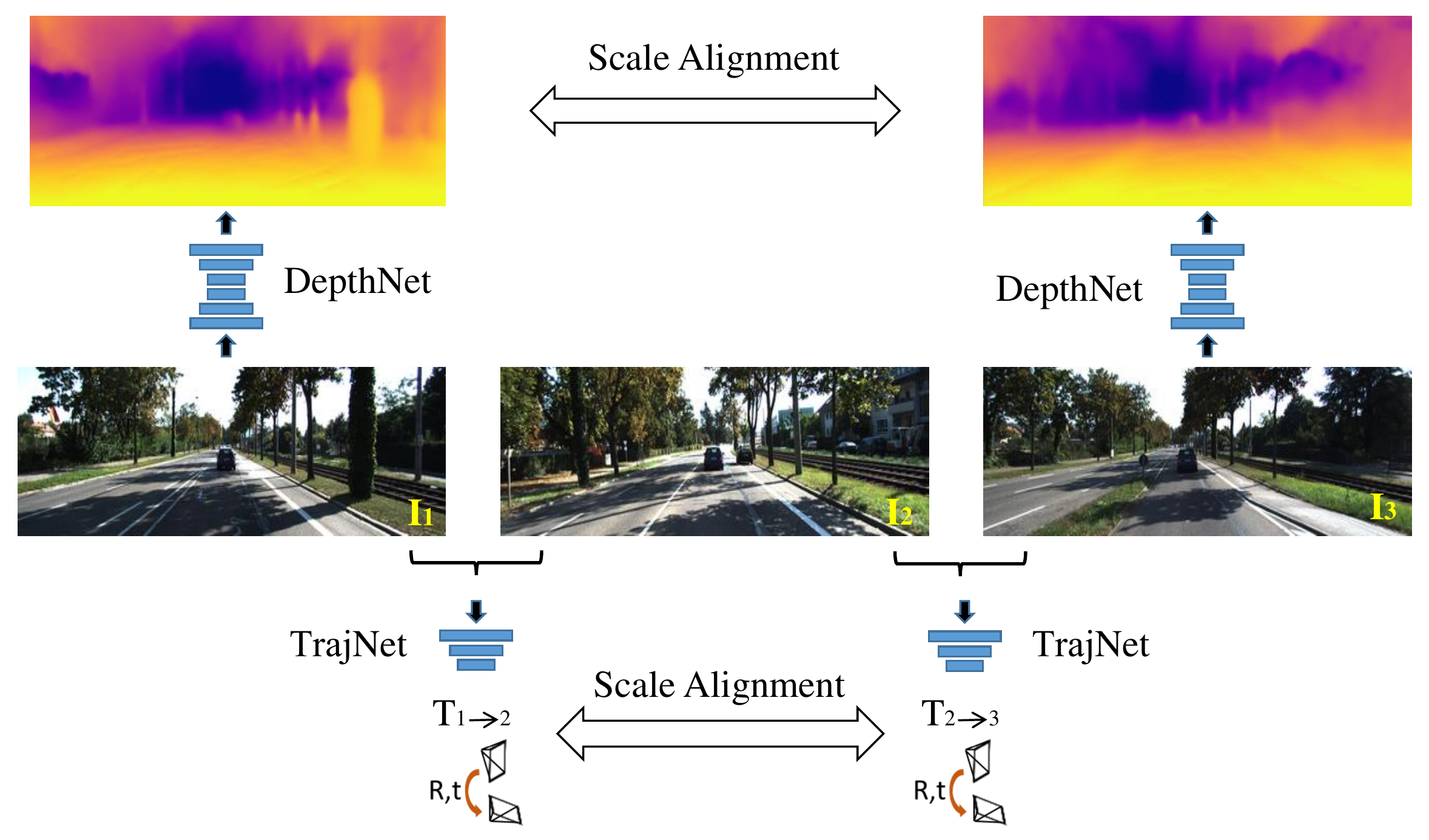}
	\caption{Scale-alignment framework in this paper. Two components are used to improve the scale-consistency of TrajNet, including the depth alignment loss $\mathcal{L}_{da}$ and the pose-to-trajectory loss $\mathcal{L}_{p2t}$. }
	\label{fig:fig3}
\end{figure}

Although the above monocular-based methods \cite{zhou2017unsupervised,yin2018geonet} achieve an accurate pose estimation between adjacent frames, their results suffer from scale-ambiguity and scale-inconsistency. The scale-inconsistency of the pose network refers to the scale of predicted poses between different adjacent frames in the same video being inconsistent. Therefore, these methods cannot accurately predict the trajectory over the long monocular video because of the per-pose scale-inconsistency. Taking the well-known monocular unsupervised methods \cite{zhou2017unsupervised,yin2018geonet} as an example, 5-frame or 3-frame short snippets are used to train their pose and depth networks.
For every snippet, the middle frame of this snippet is regarded as the target image, and its pixel-level depth map is predicted by the depth network. Meanwhile, the other frames of this snippet are taken as source images. Each snippet is sent to the pose network, and the poses between every source frame and the target image are simultaneously regressed by the pose network. Then, the view reconstruction is used to synthesize the target image from every source image independently by utilizing the predicted poses and depth map, and the differences between the real and synthesized target images are used as the main supervised signal during training.
Since the monocular sequences do not contain scale information, the trained networks suffer from scale ambiguity.
Meanwhile, because the contribution of each snippet to network training is independent, the scale between different frame snippets is inconsistent, thereby resulting in the scale-inconsistency problem.
Recovering the scale is one of the most effective ways to address scale-inconsistency. Therefore, Zhan et al.  \cite{zhan2018unsupervised} propose a novel training framework, and the pose network in \cite{zhan2018unsupervised} is trained by stereo image sequences, so that the real scale information can be extracted from the stereo image pairs during training.
Considering the lack of scale information in monocular sequences, the monocular unsupervised framework faces a major challenge in scale consistency.
Bian et al. \cite{bian2019depth} and Zhao et al. \cite{zhao2020masked} address this problem by proposing a novel geometric constraint for the monocular unsupervised framework, and the scale between different depth maps predicted by depth network is aligned by their proposed constraints. However, their proposed constraints only focus on the scale alignment of depth maps.
In this paper, we improve the scale-consistency of TrajNet by aligning both the scale information of depth maps and poses, as shown in Fig. \ref{fig:fig3}. Based on the widely used unsupervised monocular framework \cite{zhou2017unsupervised,yin2018geonet,bian2019depth,zhao2020masked}, we improve the previous view reconstruction algorithm and propose a novel pose-to-trajectory constraint to promote the trajectory prediction ability of TrajNet.

\begin{figure*}[t]
	\centering
	\includegraphics[width = 1.8\columnwidth]{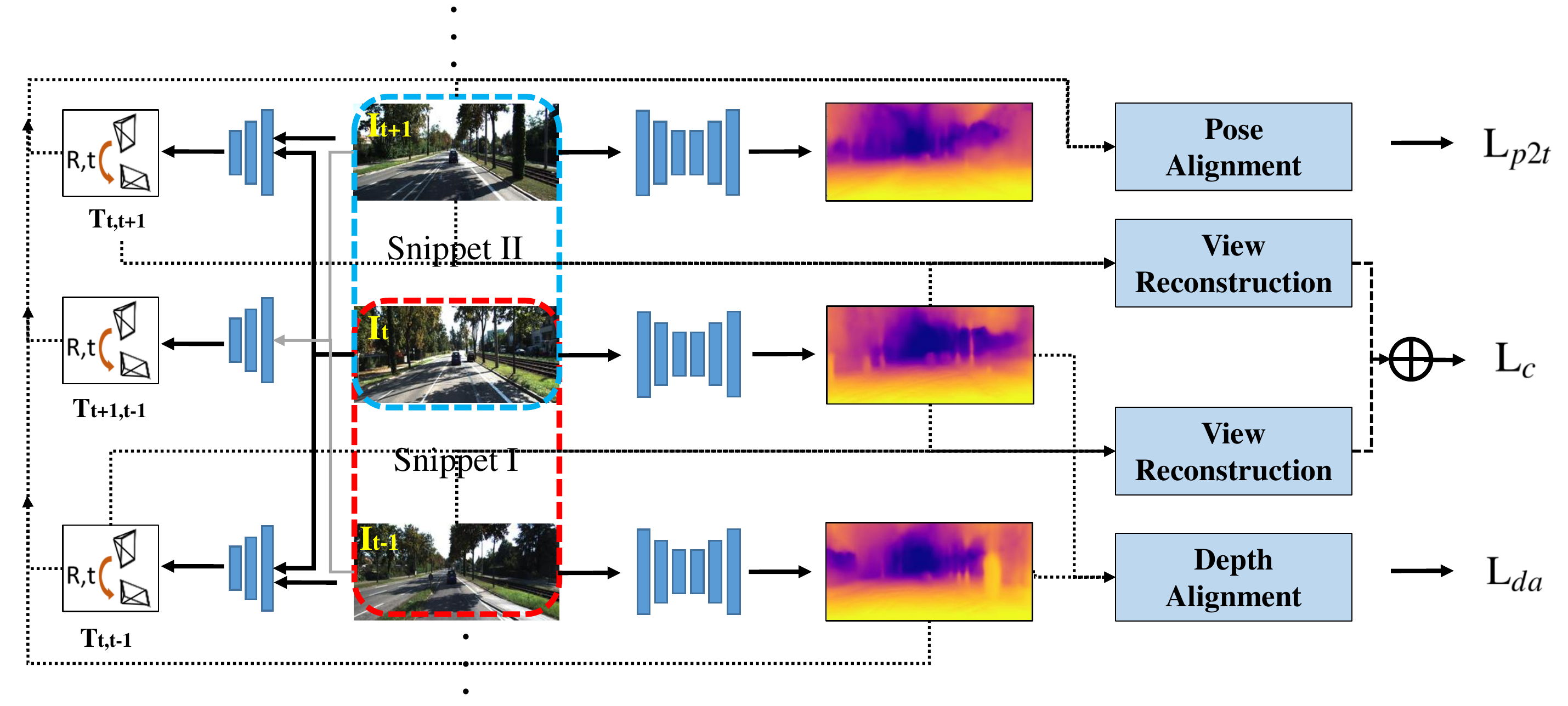}
	\caption{ The unsupervised scale-consistent network architecture. As shown in this figure, TrajNet takes the adjacent frames from monocular videos and regresses the pose between frames, and DepthNet estimates the pixel-level depth map from a single image. To improve the scale-consistency of TrajNet, we take several measures, including improving the existing framework and proposing novel constraints. The scale of different snippets, such as between snippet I and snippet II, is aligned by the view reconstruction loss $\mathrm{L}_{c}$, depth alignment loss $\mathrm{L}_{da}$ and pose-to-trajectory loss $\mathrm{L}_{p2t}$, so that TrajNet has a good scale-consistency for full trajectory prediction over long monocular videos.}
	\label{fig:fig2}
\end{figure*}

\subsection{Geometry-based framework with deep learning}

With the powerful image data processing capabilities of deep learning, many methods \cite{tang2020overview} have been proposed to improve the performance of traditional geometry-based architectures by combining deep learning frameworks. For example, Muresan et al. \cite{muresan2020stabilization} introduce the semantic information to validate the measured position of objects and improve inconsistencies between different sensor measurements.
Here, we focus on the combination of a geometry-based framework and deep learning-based pose or depth estimation. To improve the performance of geometry-based visual odometry, Tateno et al. \cite{tateno2017cnn} regard the predicted depth map as the initial guess for LSD-SLAM, and the predicted depth value is then improved by local or global optimization algorithms. Similar to \cite{tateno2017cnn}, Loo et al. \cite{loo2019cnn} introduce deep learning-based depth estimation into SVO \cite{forster2016svo} to improve the mapping of SVO. Yang et al. \cite{yang2018deep} also incorporate the depth map predicted by a deep network into DSO \cite{engel2017direct}. Since the depth network is trained on stereo image pairs, the predicted depth map contains the real scale information.  Therefore, the framework proposed in \cite{yang2018deep} can address the scale ambiguity in monocular VO. Compared with the depth network, the pose network has a better real-time performance with fewer parameters.
To improve the accuracy of depth estimation, Wang et al. \cite{wang2018Learning} combine the direct method with an unsupervised framework to provide a more accurate pose estimation; therefore, the performance of the depth network is improved. In addition, they also introduce the output of the pose network into a direct method to provide a better pose estimation between frames during the training process. However, the method proposed in \cite{wang2018Learning} only focuses on improving the depth estimation by using the direct method. More recently, Yang et al. \cite{yang2020d3vo} propose a novel unsupervised monocular depth estimation network trained on stereo videos, and both the depth, pose, and uncertainty estimated by networks are introduced into DSO \cite{engel2017direct}, thereby boosting both front-end tracking and back-end non-linear optimization. However, their method mainly focuses on the effectiveness of depth estimation on improving DSO and lacks the discussion of only introducing pose estimation. Moreover, the unsupervised framework based on stereo videos in \cite{yang2020d3vo} relies heavily on having accurate calibration between stereo cameras. Compared with \cite{yang2020d3vo}, our TrajNet is trained on monocular videos in an unsupervised manner, and we study the effectiveness of deep learning-based pose estimation in improving the initialization and tracking of DSO \cite{engel2017direct}.

\section{METHODS}

In this section, we introduce the unsupervised framework and geometric constraints for the training of TrajNet in part A, and the proposed DDSO architecture is described in part B.

\subsection{Unsupervised framework}

In this subsection, we first introduce the basic unsupervised loss functions widely used in \cite{zhou2017unsupervised,yin2018geonet,bian2019depth,zhao2020masked}, including the view reconstruction constraint $\mathrm{L}_{c}$, the smoothness constraint $\mathrm{L}_{smooth}$ and the depth alignment constraint $\mathrm{L}_{da}$. To further improve the scale-consistency of TrajNet, we improve the view reconstruction constraint $\mathrm{L}_{c}$ to constrain the scale-inconsistency between input snippets. Furthermore, we propose a novel pose-to-trajectory constraint $\mathrm{L}_{p2t}$ to align the scale factor between poses.

\textbf{Total unsupervised framework:} Instead of using the expensive ground truth for training TrajNet, a general unsupervised framework is considered to effectively train TrajNet in this study. TrajNet is trained by monocular RGB sequences composed of a target frame $I_{t}$ and its adjacent frame $I_{t-1}$ and regresses the 6-DoF transformation $\hat{T}_{t,\,t-1}$ of them. Simultaneously, a depth map $\hat{D}_{t}$ of the target frame is generated by the depth network (called DepthNet). The geometry constraints between the outputs of two deep models serve as a training monitor that helps the model learn the geometric relations between adjacent frames, as shown in Fig. \ref{fig:fig2}.
The key supervisory signal for TrajNet comes from the view reconstruction loss $\mathrm{L}_{c}$, smoothness loss $\mathrm{L}_{smooth}$, depth alignment loss $\mathrm{L}_{da}$ and pose-to-trajectory loss $\mathrm{L}_{p2t}$:
\begin{equation}
\mathrm{L} = \mathrm{L}_{c} +\alpha \mathrm{L}_{smooth} + \beta \mathrm{L}_{da} + \gamma \mathrm{L}_{p2t}, \label{eq:1}
\end{equation}
where $\alpha$, $\beta$ and $\gamma$ are balance factors. 

\textbf{View reconstruction constraint:} During training, two consecutive frames including the target frame $I_{t}$ and source frame $I_{t-1}$ are concatenated along the channel dimension and fed into TrajNet to regress the 6-DoF camera pose $\hat{T}_{t \to t-1}$. DepthNet takes a single target frame $I_{t}$ as input and outputs the per-pixel depth prediction $\hat{D}_{t}$. As indicated in Eq. (\ref{eq:2}), we can obtain the pixel correspondence between two frames based on the geometric projection, which is similar to \cite{zhou2017unsupervised,bian2019depth}:
\begin{equation}
p_{t-1} \sim K \hat{T}_{t \to t-1} \hat{D}_{t}(p_{t})K^{-1}p_{t}, \label{eq:2}
\end{equation}
where $K$ is the camera intrinsics matrix. Note that $p_{t}$ is continuous on the image while the projection is discrete. To warp the source frame $I_{t-1}$ to target frame $I_{t}$ and obtain a continuous smooth reconstruction frame $\hat{I}_{t-1}$, we use the differentiable bilinear interpolation mechanism. We assume that the scenes used in training are static and adopt a robust image similarity loss \cite{godard2017unsupervised} for the consistency of the two views $I_{t}$, $\hat{I}_{t-1}$:
\begin{equation}
\mathrm{L}^{t \to t-1}_{c} = \delta_{1} \dfrac{1-SSIM(I_{t},\hat{I}_{t-1})}{2} + (1-\delta_{1}) ||I_{t}-\hat{I}_{t-1}||_{1}, \label{eq:3}
\end{equation}
where $SSIM(I_{t},\hat{I}_{t-1})$ stands for the structural similarity \cite{Zhou2004Image} between $I_{t}$ and $\hat{I}_{t-1}$.

 To enhance the scale-consistency between different snippets, we combine the current frame $I_{t}$ and the next frame $I_{t+1}$ as a novel and independent snippet, and the pose between $I_{t}$ and $I_{t+1}$ is regressed by TrajNet. Therefore, based on the predicted depth map of $I_{t}$, so that the view reconstruction constraint between $I_{t}$ and $I_{t+1}$ is calculated:
\begin{equation}
\mathrm{L}^{t \to t+1}_{c} = \delta_{1} \dfrac{1-SSIM(I_{t},\hat{I}_{t+1})}{2} + (1-\delta_{1}) ||I_{t}-\hat{I}_{t+1}||_{1}, \label{eq:4}
\end{equation}
where $\hat{I}_{t+1}$ stands for the image synthesized from ${I}_{t+1}$ by view reconstruction. Therefore, the final view reconstruction constraint is represented as follows:
\begin{equation}
\mathrm{L}_{c} = \mathrm{L}^{t \to t-1}_{c} + \mathrm{L}^{t \to t+1}_{c}. \label{eq:5}
\end{equation}
Since the view reconstruction errors of different snippets are calculated based on the same depth map, the scale of poses is aligned with that of the depth map, so that the consistency of TrajNet between different image snippets is improved. This is different with previous works \cite{zhou2017unsupervised,yin2018geonet,bian2019depth,zhao2020masked} that only consider the loss calculations within a snippet or depth alignment between snippets.

\textbf{Smoothness constraint:} This loss term is used to promote the representation of geometric details. There are many planes in the scenes, and the depth of adjacent pixels in the same plane presents gradient changes. Therefore, similar to \cite{godard2017unsupervised}, this paper uses the edge-aware smoothness:
\begin{equation}
\mathrm{L}_{smooth} = |\partial_{x} d_{t}^{*}|e^{\partial_{x} I_{t}} + |\partial_{y} d_{t}^{*}|e^{\partial_{y} I_{t}}, \label{eq:6}
\end{equation}
where $d_{t}^{*}=d_{t}/\hat{d_{t}}$ represents the mean-normalized inverse depth.

\textbf{Depth alignment constraint:} 
Following Bian \textit{et al.} \cite{bian2019depth} and Zhao \textit{et al.} \cite{zhao2020masked}, we promote the scale-consistency of DepthNet by aligning the scale of adjacent predicted depth maps. Because TrajNet is tightly coupled with DepthNet, the scale-consistency of TrajNet is also improved by this constraint. The depth alignment constraint is formulated as follows:
\begin{equation}
\mathrm{L}_{da} =  ||D^{t}_{s}-\hat{D}^{s}_{s}||_{1},  \label{eq:7}
\end{equation}
where,
\begin{equation}
D^{t}_{s}(p_{s}) \sim K \hat{T}_{t \to s} \hat{D}_{t}(p_{t})K^{-1}p_{t}. \label{eq:8}
\end{equation}
$D^{t}_{s}(p_{s})$ is computed by the projection algorithm shown in Eq. (\ref{eq:6}), and it has the same scale information as $\hat{D}_{t}$. $\hat{D}_{t}$ stands for the predicted depth map of the target image $I_{t}$. $\hat{D}_{s}$ is the predicted depth map of source image $I_{s}$. $\hat{D}^{s}_{s}$ is reconstructed from $\hat{D}_{s}$ by the warping algorithm, which is similar to the view reconstruction process in \cite{zhou2017unsupervised}, and it contains the same scale information as $\hat{D}_{s}$. Then, SSIM loss is adopted for the consistency of $D^{t}_{s}$ and $\hat{D}^{s}_{s}$ so that the scale between different depth maps is aligned, as shown in Fig. \ref{fig:fig3}.

\begin{figure*}[!htbp]
	\centering
	\includegraphics[width = 1.8\columnwidth]{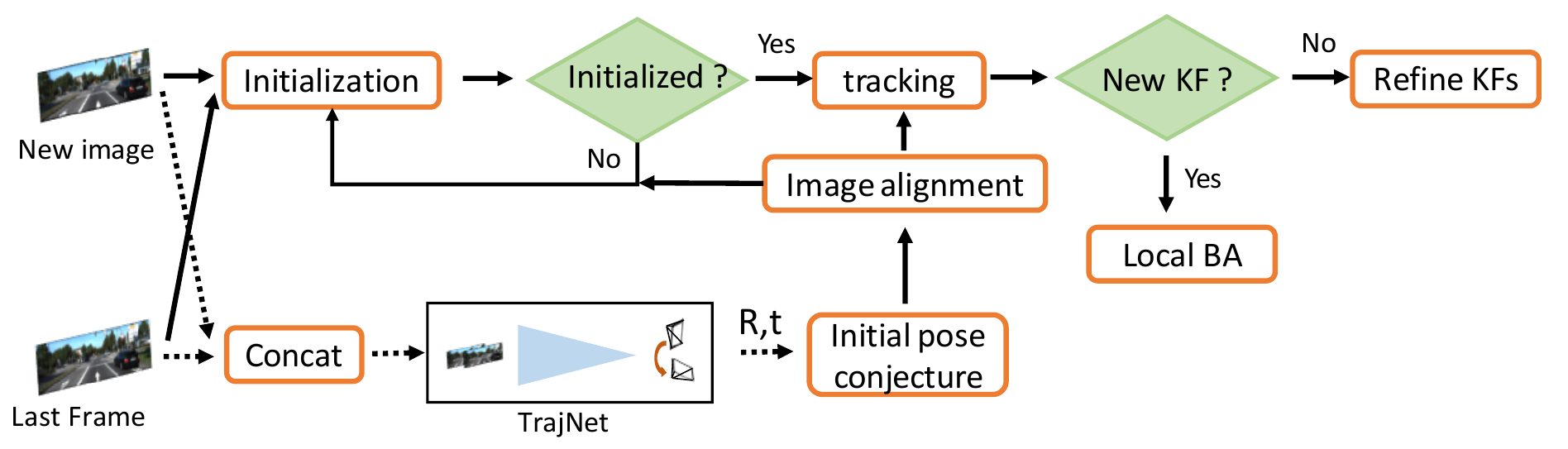}
	\caption{The DDSO pipeline. This work augments the DSO \cite{engel2017direct} framework with the pose prediction module (TrajNet). Every new frame is fed into the proposed TrajNet with the last frame to regress a relative pose estimation. The predicted pose is used to improve initialization and tracking in DSO \cite{engel2017direct}. }
	\label{fig:fig4}
\end{figure*}

\textbf{Pose-to-trajectory constraint:} To further improve the trajectory generation ability of TrajNet, a novel pose-to-trajectory constraint is proposed in this paper. The scale-inconsistency of different predicted poses ($\hat{T}_{1 \to 2}$, $\hat{T}_{2 \to 3}$, $\hat{T}_{1 \to 3}$) between adjacent frames ($I_{1}$, $I_{2}$, $I_{3}$) is utilized to improve the scale consistency between poses. If the scale information is consistent and the pose estimation is accurate, the relative poses $\hat{T}_{1 \to 2}$, $\hat{T}_{2 \to 3}$, $\hat{T}_{1 \to 3}$ satisfy the following constraint:
 \begin{equation}
 \hat{T}_{1 \to 3} = \hat{T}^{c}_{1 \to 3}, \label{eq:9}
 \end{equation}
 where:
 \begin{equation}
 \hat{T}^{c}_{1 \to 3} = \hat{T}_{1 \to 2} \hat{T}_{2 \to 3}. \label{eq:10}
 \end{equation}
 To satisfy the above constraint, the error between the image $\hat{I}_{3}$ reconstructed based on $\hat{T}_{1 \to 3}$ and the image $\hat{I}^{c}_{3}$ reconstructed based on $\hat{T}^{c}_{1 \to 3}$ is calculated to constrain the scale and accuracy of poses:
 \begin{equation}
 \mathrm{L}_{p2t} =  ||\hat{I}_{3}-\hat{I}^{c}_{3}||_{1}. \label{eq:11}
 \end{equation}

\subsection{Deep Sparse Visual Odometry}

To address the drawbacks of DSO \cite{engel2017direct}, we propose a novel framework, called DDSO, which incorporates TrajNet with DSO \cite{engel2017direct} to improve the initialization and tracking of DSO \cite{engel2017direct}. In this subsection, we analyse the drawbacks of DSO \cite{engel2017direct} and introduce the proposed DDSO framework.

\textbf{DSO:} DSO \cite{engel2017direct} is a keyframe-based approach, where 5-7 keyframes are maintained in the sliding window and their parameters are jointly optimized by minimizing photometric errors in the current window. The total photometric error $\textbf{E}_{total}$ (Eq. (\ref{eq:14})) of the sliding window is optimized by the Gauss-Newton algorithm and used to calculate the relative transformation $\textbf{T}_{ij}$.


\begin{equation}
\textbf{p}^{\prime} = \pi_{c}(\textbf{R} \pi^{-1}_{c}( \textbf{p}, d_{p}) + t ), \qquad \left[
\begin{matrix}
R_{ij} & t_{ij}  \\
0 & 1
\end{matrix}
\right] = \textbf{T}_{ij},  \label{eq:12}
\end{equation}

\begin{equation}
\textbf{E}_{\textbf{p}j} : =\sum_{p\in\mathit{N}_{\textbf{p}}} w_{p} \lVert I_{j}[\textbf{p}^{\prime}] - I_{i}[\textbf{p}] \rVert_{\gamma}, \label{eq:13}
\end{equation}

\begin{equation}
\textbf{E}_{total} : =\sum_{i\in\mathbb{F}} \sum_{p\in\mathbb{P}_{i}} \sum_{j\in\mathnormal{obs(p)}} \textbf{E}_{\textbf{p}j}, \label{eq:14}
\end{equation}
where $\pi_{c}$ is the projection function: $\mathbb{R}^{3} \to \omega $ while $\pi^{-1}_{c}$ is back-projection. $\textbf{p}^{\prime}$ stands for the projected point position of \textbf{p} with inverse depth $d_{p}$. $\textbf{T}_{ij}$ is the transformation between two related frames $I_{i}$ and $I_{j}$. $\mathbb{F}$ is a collection of frames in the sliding window, and $\mathbb{P}_{i}$ refers to the points in frame $i$.  $\mathnormal{obs(p)}$ means that the points are visible in the current frame.

Since the whole process can be regarded as a nonlinear optimization problem, an initial transformation should be given and iteratively optimized by the Gauss-Newton method. Therefore, the initial transformation, especially orientation, is very important for the whole tracking process. During tracking, a constant motion model is applied for initializing the relative transformation between the current frame and last key-frame in DSO \cite{engel2017direct}, as shown in Eq. (\ref{eq:15}) and Eq. (\ref{eq:16}), assuming that the motion $T_{t,\,t-1}$ between the current frame $I_{t}$ and last frame $I_{t-1}$ is the same as the previous one $T_{t-1,\,t-2}$:

\begin{equation}
\textbf{T}_{t,\,t-1} = \textbf{T}_{t-1,\,t-2} = \textbf{T}_{t-1,\,w} \cdot \textbf{T}^{-1}_{t-2,\,w}, \label{eq:15}
\end{equation}
\begin{equation}
\textbf{T}_{t,\,kf} = \textbf{T}_{t,\,t-1} \cdot \textbf{T}_{t-1,\,kf} =  \textbf{T}_{t,\,t-1} \cdot \textbf{T}_{t-1,\,w} \cdot \textbf{T}^{-1}_{kf,w}, \label{eq:16}
\end{equation}
where $\textbf{T}_{t-1,\,w},\textbf{T}_{t-2,\,w},\textbf{T}_{kf,\,w}$ are the poses of $I_{t-1}, I_{t-2}, I_{kf} $ in world coordinate system.

\textbf{DDSO:} Considering that it is not reliable to use only the initial transformation provided by the constant motion model, DSO \cite{engel2017direct} attempts to recover the tracking process by initializing the other 3 motion models and 27 different small rotations when the image alignment algorithm fails, which is complex and time consuming. Since there is no motion information as a priori during the initialization process, the transformation is initialized to the identity matrix, and the inverse depth of the point is initialized to 1.0. In this process, the initial value of optimization is meaningless, resulting in inaccurate results and even initialization failure.

Therefore, we utilize TrajNet to provide an accurate initial transformation, especially orientation, for the initialization and tracking processes in this paper. During the initialization process, the identity matrix is replaced by the poses between adjacent frames predicted by TrajNet. Similarly, the constant motion model as well as other motion models in the tracking process are also replaced by the predicted poses:
\begin{equation}
\textbf{T}_{t,t-1} = \textbf{TrajNet}(I_{t},I_{t-1}), \label{eq:17}
\end{equation}
With the help of TrajNet, a better pose estimation can be regarded as a better guide for initialization and tracking. As shown in Fig. \ref{fig:fig4}, DDSO builds on the monocular DSO \cite{engel2017direct}, and the pose predictions provided by our TrajNet are used to improve DSO in both the initialization and tracking processes. When a new frame arrives, a relative transformation $\textbf{T}_{t,\,t-1}$ is regressed by the pre-trained TrajNet from the current frame $I_{t}$ and the last frame $I_{t-1}$. The pose estimated by TrajNet is regarded as the initial value of the image alignment algorithm. Due to a more accurate initial value provided by TrajNet for the nonlinear optimization process, the robustness of DSO tracking is improved.

\section{Experiment}

We evaluate TrajNet as well as DDSO against the state-of-the-art methods on the publicly available KITTI dataset \cite{geiger2013vision}.

\subsection{TrajNet}

\textbf{Training framework:}
TrajNet is trained in an unsupervised manner, and the unsupervised framework consists of two modules, TrajNet and DepthNet. TrajNet takes the adjacent two frames and regresses the 6-DoF pose between frames. DepthNet predicts the pixel-level depth map from a single image. Both the poses and depth maps are predicted in an end-to-end manner. For fairness in comparison, the network architectures of TrajNet and DepthNet in this paper remain the same as the previous methods \cite{zhou2017unsupervised,yin2018geonet,bian2019depth,zhao2020masked}. The main difference lies in the design of the loss function, which is important in the unsupervised framework.

\textbf{Training detail:} We implement the architecture with the Tensorflow framework and train on an NVIDIA RTX 2080 Ti GPU. The loss weights are set to $\alpha = 0.5$, $\beta = 0.5$, $\gamma = 0.5$, and $\delta_{1}=0.85$. 
Networks are trained by the ADAM optimizer with $\beta_{1} = 0.9$. During training, the resolution of input images is adjusted to $128\times416$. The learning rate is initialized as 0.0002 and the mini-batch is set as 8. The training converges after approximately 150K iterations. The KITTI sequences 00-08 of the KITTI odometry dataset \cite{geiger2013vision} are applied to train TrajNet, and sequences 09-10 are used to evaluate TrajNet. For fairness, the usage of training and testing sets is the same as in previous works \cite{zhou2017unsupervised,yin2018geonet,bian2019depth,zhao2020masked}.

\begin{figure}[t]
	\centering
	\subfigure[Predicted trajectory on Seq. 09]{
		\includegraphics[width = 0.45\columnwidth]{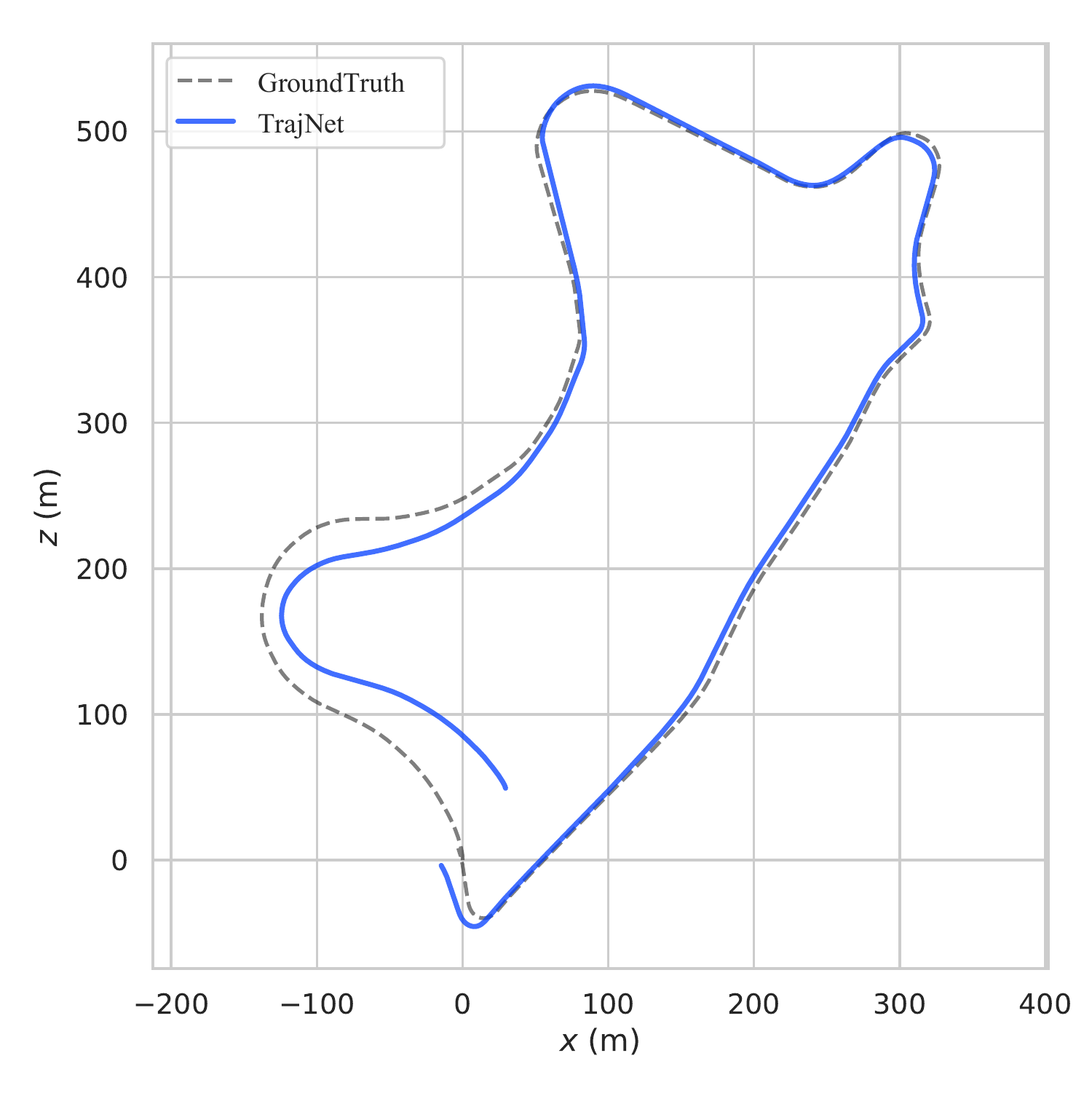}
	}
	\subfigure[Predicted trajectory on Seq. 10]{
		\includegraphics[width = 0.45\columnwidth]{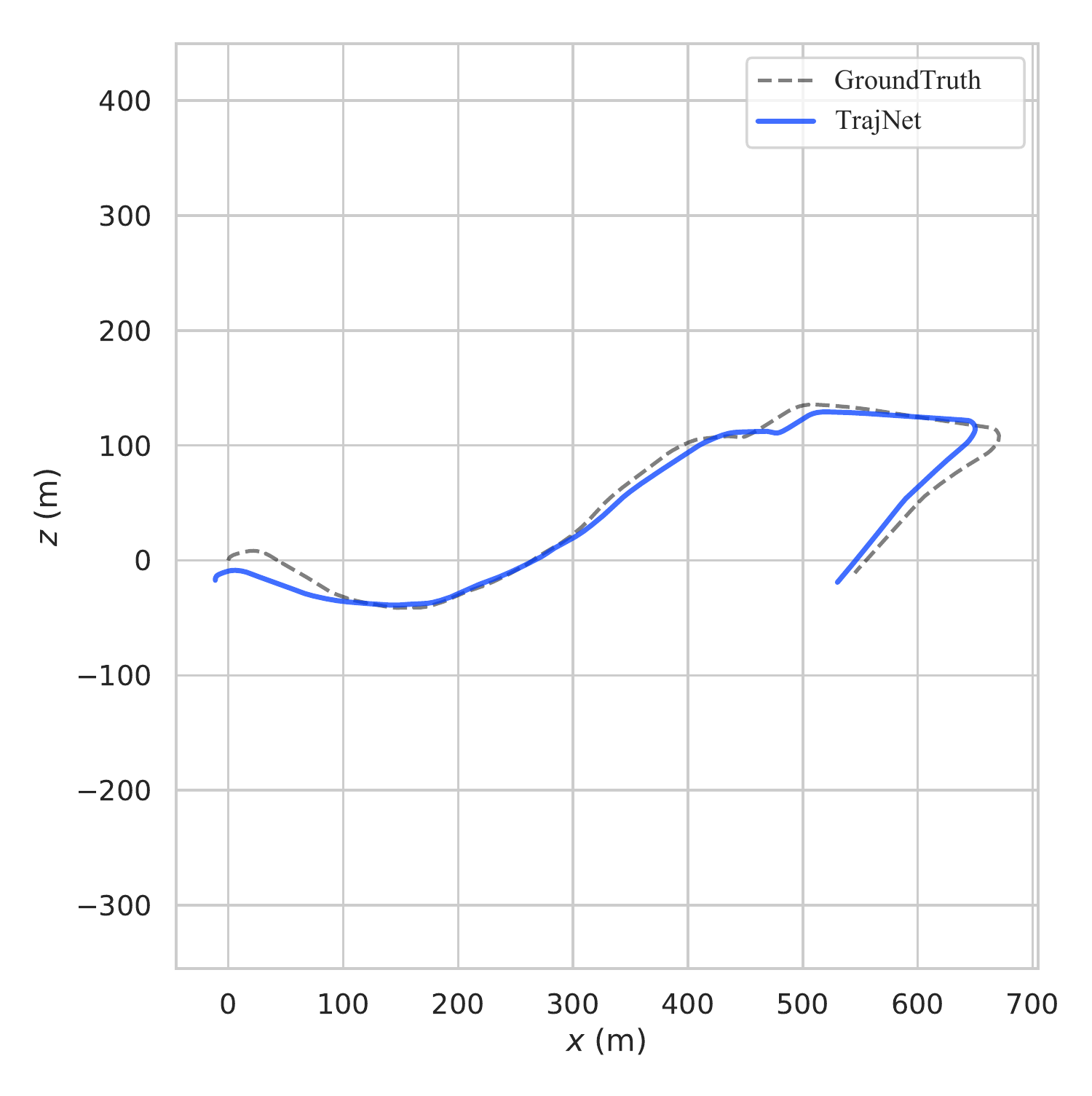}
	}
	\caption{ Trajectories predicted by TrajNet on KITTI odometry sequences 09-10. As shown in (a) and (b), TrajNet has the ability to generate the full trajectory of a long monocular video.}
	\label{fig:fig5}
\end{figure}

\begin{figure}[t]
	\centering
	\subfigure[Trajectories on Seq. 09]{
		\includegraphics[width = 0.9\columnwidth]{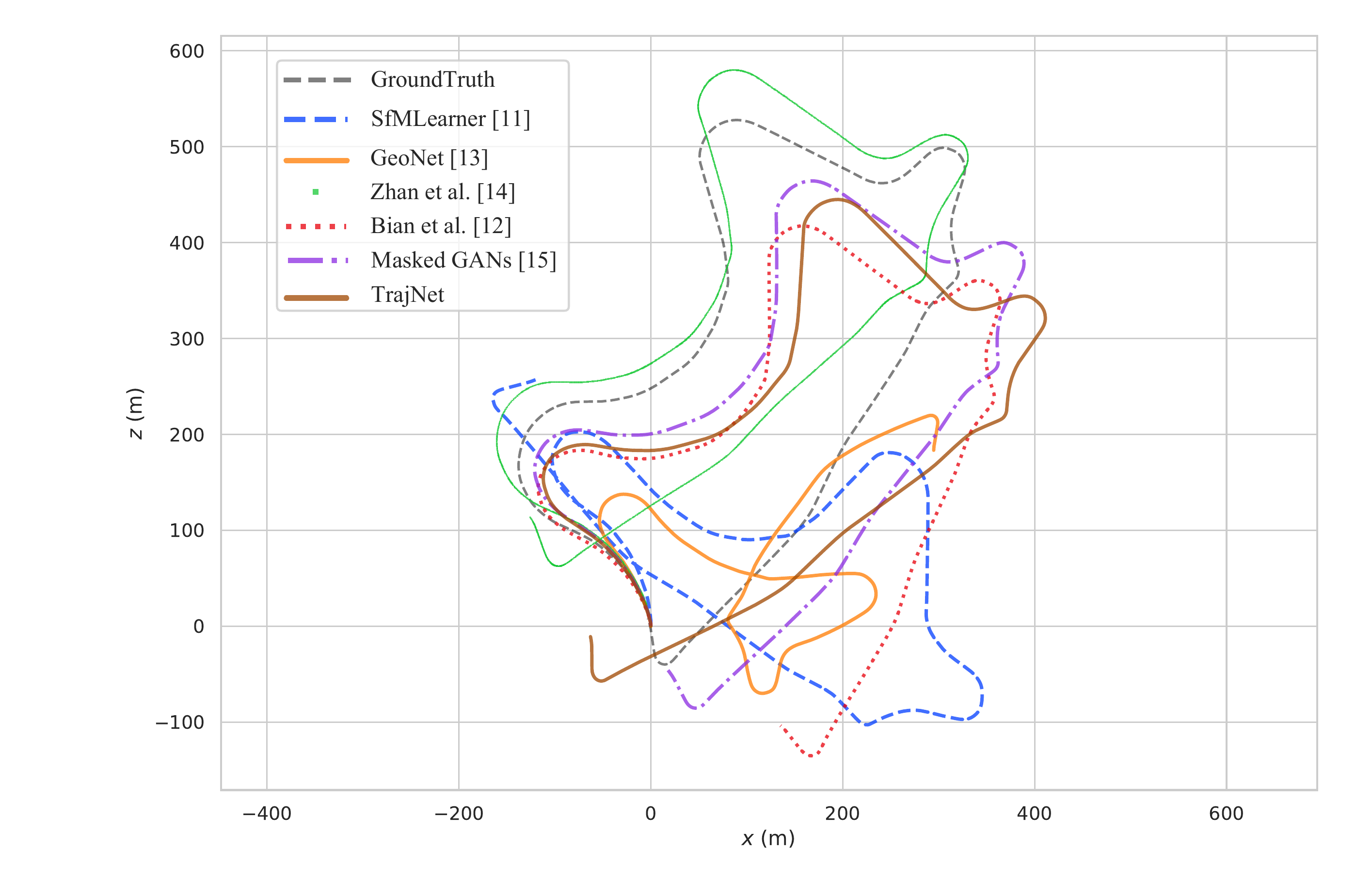}
	}
	\subfigure[Trajectories on Seq. 10]{
		\includegraphics[width = 0.9\columnwidth]{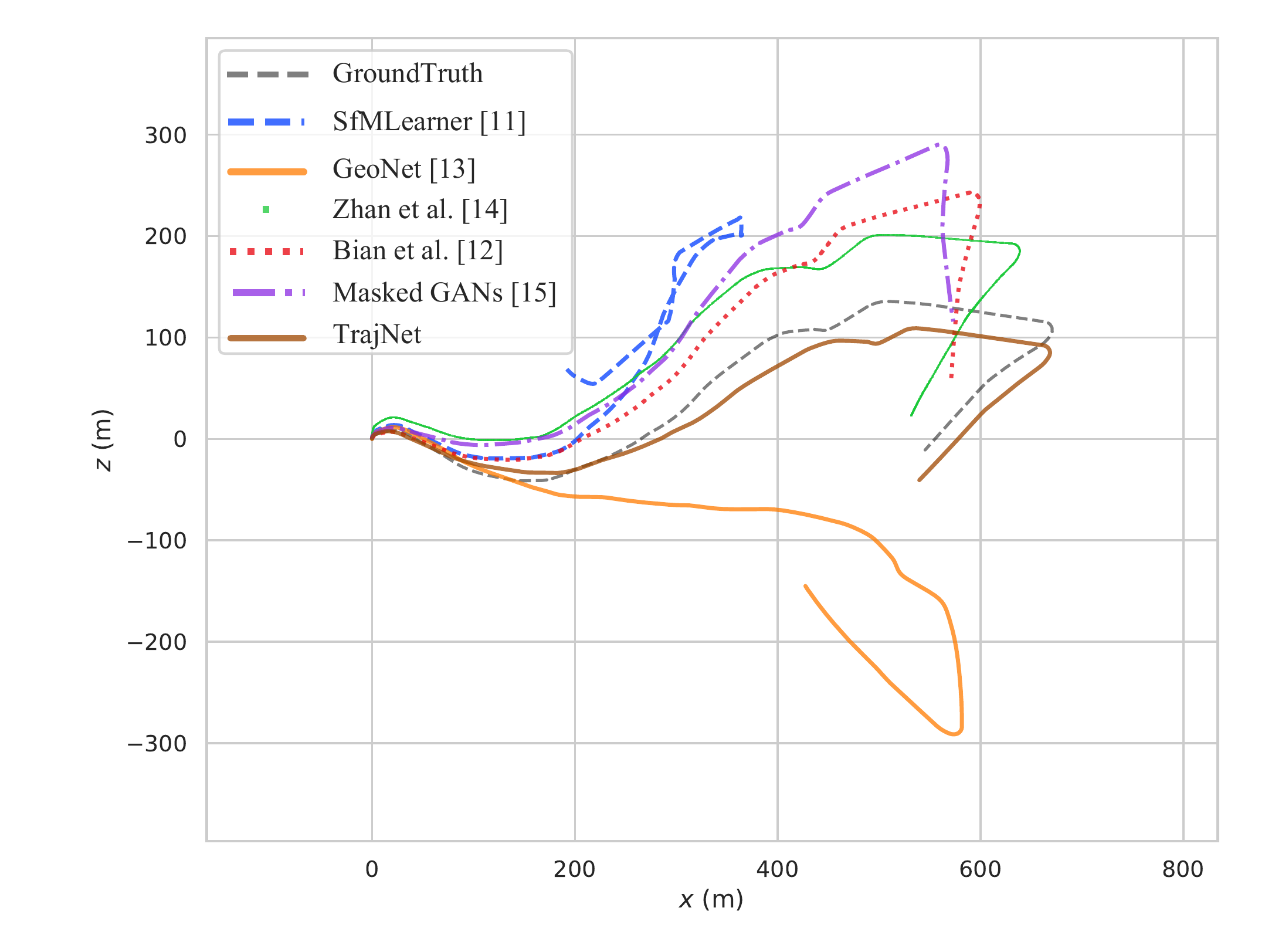}
	}
	\caption{ Comparison with different deep learning-based VO methods. From the predicted trajectories, the results of TrajNet show outstanding odometry estimation in terms of both position and orientation.}
	\label{fig:fig6}
\end{figure}

\textbf{Ablation study:} In this section, we first conduct a series of ablation experiments to validate the efficacy of the proposed framework in full trajectory prediction. As shown in Table \ref{Tab01}, the framework is trained by different loss functions in an unsupervised manner. ``$\mathrm{L}_{c}$'' refers to the view reconstruction loss widely used in previous works \cite{zhou2017unsupervised,yin2018geonet} and improved in this paper for scale consistency; ``$\mathrm{L}_{smooth}$'' stands for the smoothness loss widely used in previous works \cite{yin2018geonet,bian2019depth}. $\mathrm{L}_{da}$ denotes the depth alignment constraint similar to \cite{bian2019depth,zhao2020masked}, and $\mathrm{L}_{p2t}$ represents the novel constraint proposed in this paper for full trajectory generation. As shown in Table \ref{Tab01}, because of optimization of the view reconstruction constraint, TrajNet trained by the basic loss function has good scale consistency.
In addition, with the help of these two scale-consistent constraints, TrajNet has the ability to accurately predict the full trajectory of long monocular video, especially in the orientation estimation. Moreover, comparing lines 3 and 2 in Table \ref{Tab01} the novel proposed pose-to-trajectory constraint $\mathrm{L}_{p2t}$ effectively improves the accuracy of TrajNet in both translation and orientation estimation.

\begin{table}[t]
    \huge
	
	\scriptsize
	
	\centering
	
	\caption{ \textit{Results of ablation experiments on KITTI odometry dataset \cite{geiger2013vision}. The average translation error ``$t_{err}$'' and rotation error ``$r_{err}$'' are reported.}}
	
	\label{Tab01}
	\resizebox{\columnwidth}{!}{
		\begin{tabular}{c|cc|cc}
			
			\toprule
			\multicolumn{1}{c}{}& \multicolumn{2}{c}{Seq. 09} & \multicolumn{2}{c}{Seq. 10}  \\
			\cmidrule(r){2-3} \cmidrule(r){4-5}
			
			Loss function  &$t_{err}(\%)$	&$r_{err}(^{\circ}/100m)$ &$t_{err}(\%)$ &$r_{err}(^{\circ}/100m)$ 	 \\
			\hline
			$\mathrm{L}_{c}$ + $\mathrm{L}_{smooth}$ 		&  9.42  	&  3.55  &  10.98   &  4.98	 \\		
			$\mathrm{L}_{c}$ + $\mathrm{L}_{smooth}$ + $\mathrm{L}_{da}$  & 11.55  &  3.43  &11.70  & 4.13 \\
            $\mathrm{L}_{c}$ + $\mathrm{L}_{smooth}$ + $\mathrm{L}_{da}$ + $\mathrm{L}_{p2t}$  &  \textbf{7.40}& \textbf{2.21} & \textbf{10.28} &\textbf{2.82}\\
			\bottomrule
			
		\end{tabular}
	}
\end{table}

\begin{table}[h]
	\LARGE
	\scriptsize
	
	\centering
	
	\caption{ \textit{Visual odometry results on KITTI odometry dataset \cite{geiger2013vision}. \textbf{Sup.} refers to the source of supervisory signal. \textbf{M.} and \textbf{S.} stand for the ``monocular sequences'' and ``stereo image pairs'' respectively.}}
	
	\label{Tab02}
	\resizebox{\columnwidth}{!}{
		\begin{tabular}{c|c|cc|cc}
			
			\toprule
			\multicolumn{2}{c}{}& \multicolumn{2}{c}{Seq. 09} & \multicolumn{2}{c}{Seq. 10}  \\
			\cmidrule(r){3-4} \cmidrule(r){5-6}
			
			Method  & Sup.	&  $t_{err}(\%)$		&  $r_{err}(^{\circ}/100m)$      &  $t_{err}(\%)$    &   $r_{err}(^{\circ}/100m)$ 	 \\
			\hline
			SfMLearner \cite{zhou2017unsupervised} &M.	&  17.84   	&  6.78  &  37.91   &   17.78 	 \\
			GeoNet 	\cite{yin2018geonet}		&M.	&  41.47    &  13.14 &  32.74   &   13.12	 \\
			Zhan \textit{et al.} \cite{zhan2018unsupervised} &S.	&  11.93    &  3.91  &  12.45   & 3.46 \\
			Bian \textit{et al.}	\cite{bian2019depth} 		&M.	&  11.2    	&  3.35  &  10.1 & 4.96 \\
			Wang \textit{et al.} \cite{wang2019recurrent} 		&M.	&  9.88   	&  3.40  &  12.24   &  5.20	 \\	
            Struct2depth \cite{casser2019unsupervised}		    &M.	&  10.2     &  2.64  &29.0  &   4.28 	 \\	
			Masked GANs	\cite{zhao2020masked}					&M.	&  8.71     &  3.10  &\textbf{9.63}  &  3.42 	 \\
            TrajNet(ours)                                       &M.  &  \textbf{7.40}& \textbf{2.21} & 10.28 &\textbf{2.82}\\
			\hline
            DDSO(ours)                 & -  &  15.02 & \textbf{0.20} & \textbf{8.76} &\textbf{0.66}\\
			\bottomrule
			
		\end{tabular}
	}
\end{table}

\begin{figure*}[htbp]
	\centering
	\subfigure[Trajectory generated by DDSO on Seq. 07]{
		\includegraphics[width = 0.47\columnwidth]{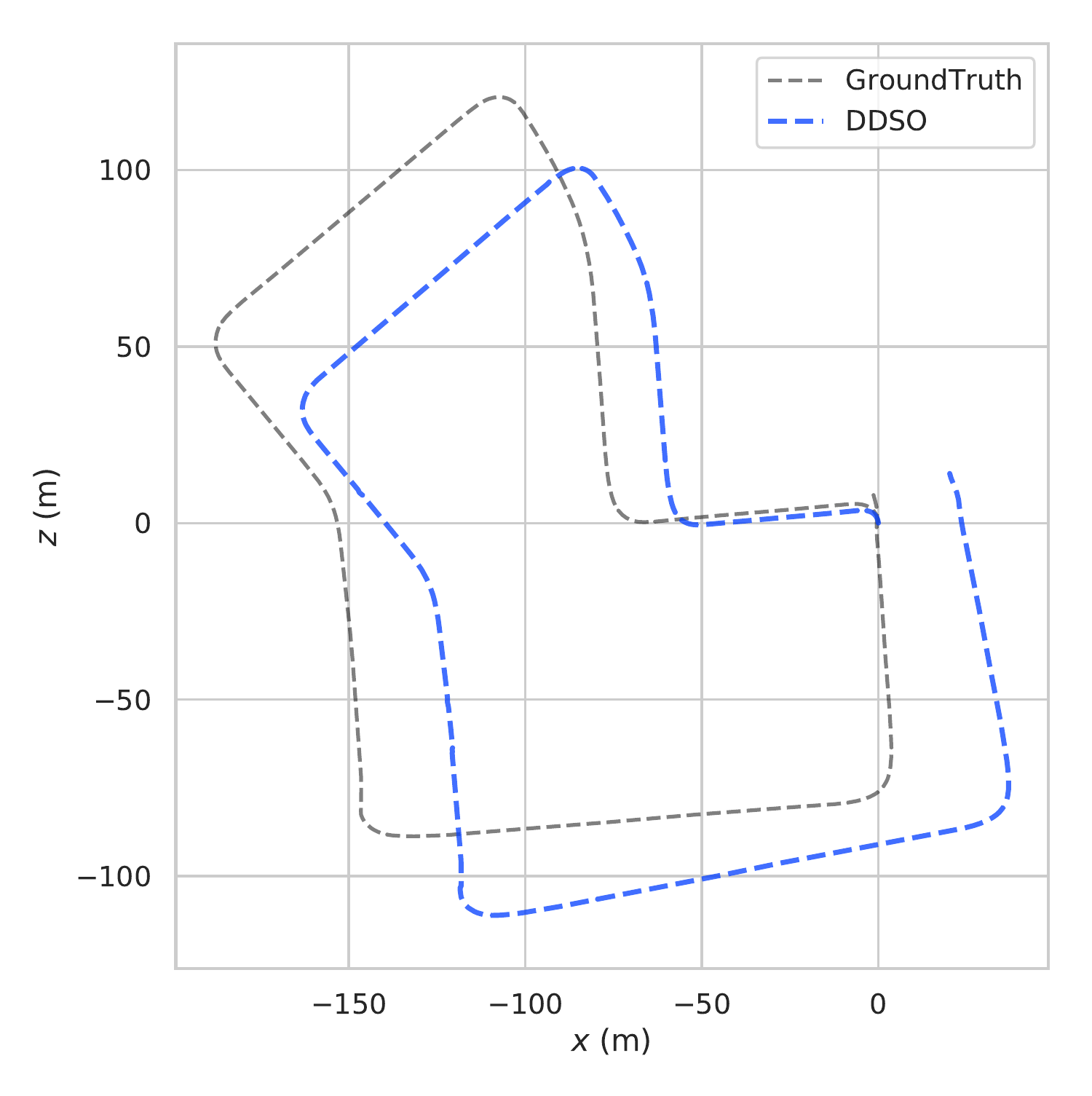}
	}
	\subfigure[The point cloud on Seq. 07]{
		\includegraphics[width = 0.47\columnwidth]{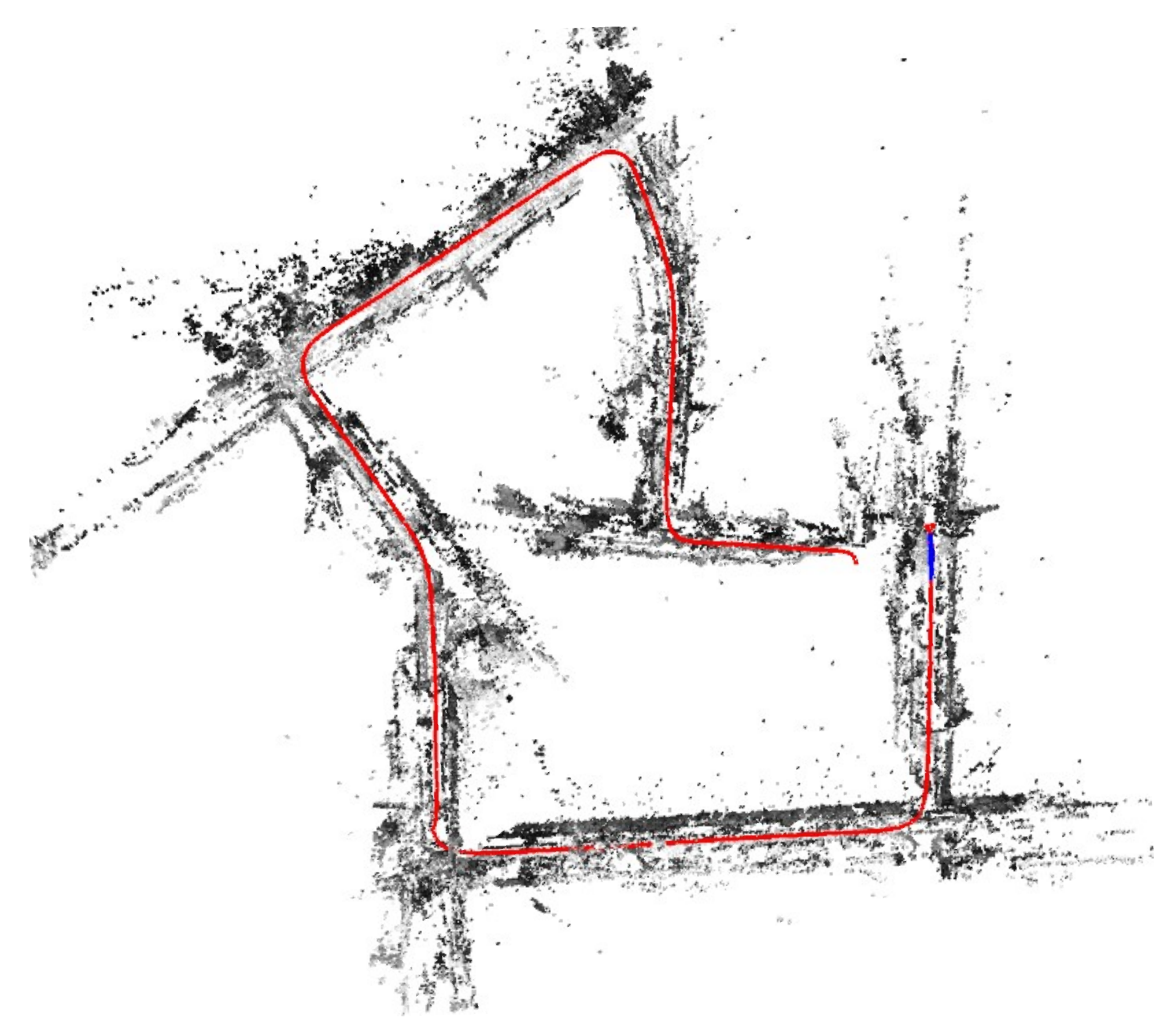}
	}
	\subfigure[Trajectory generated by DDSO on Seq. 08]{
		\includegraphics[width = 0.47\columnwidth]{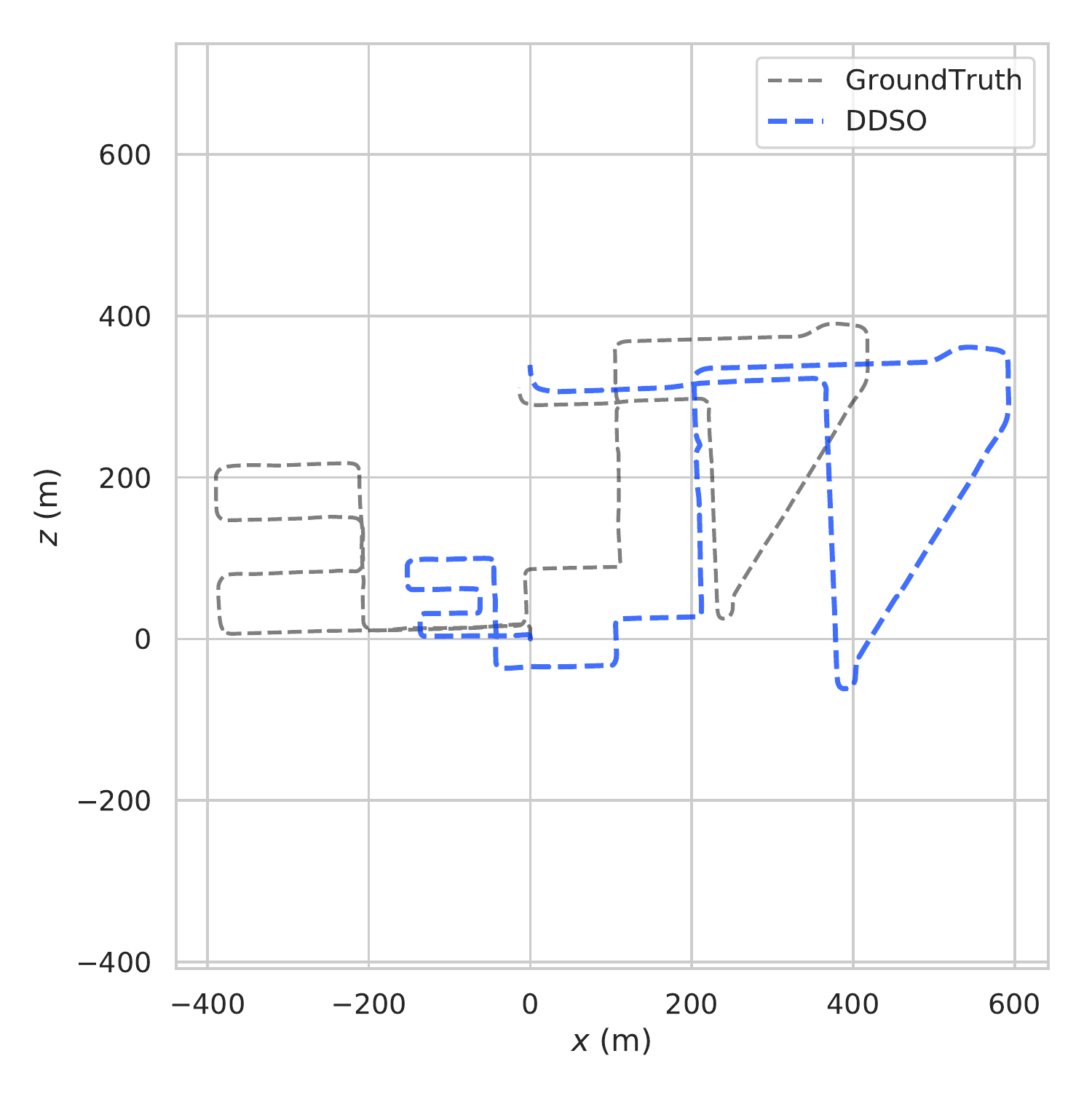}
	}
	\subfigure[The point cloud on Seq. 08]{
		\includegraphics[width = 0.47\columnwidth]{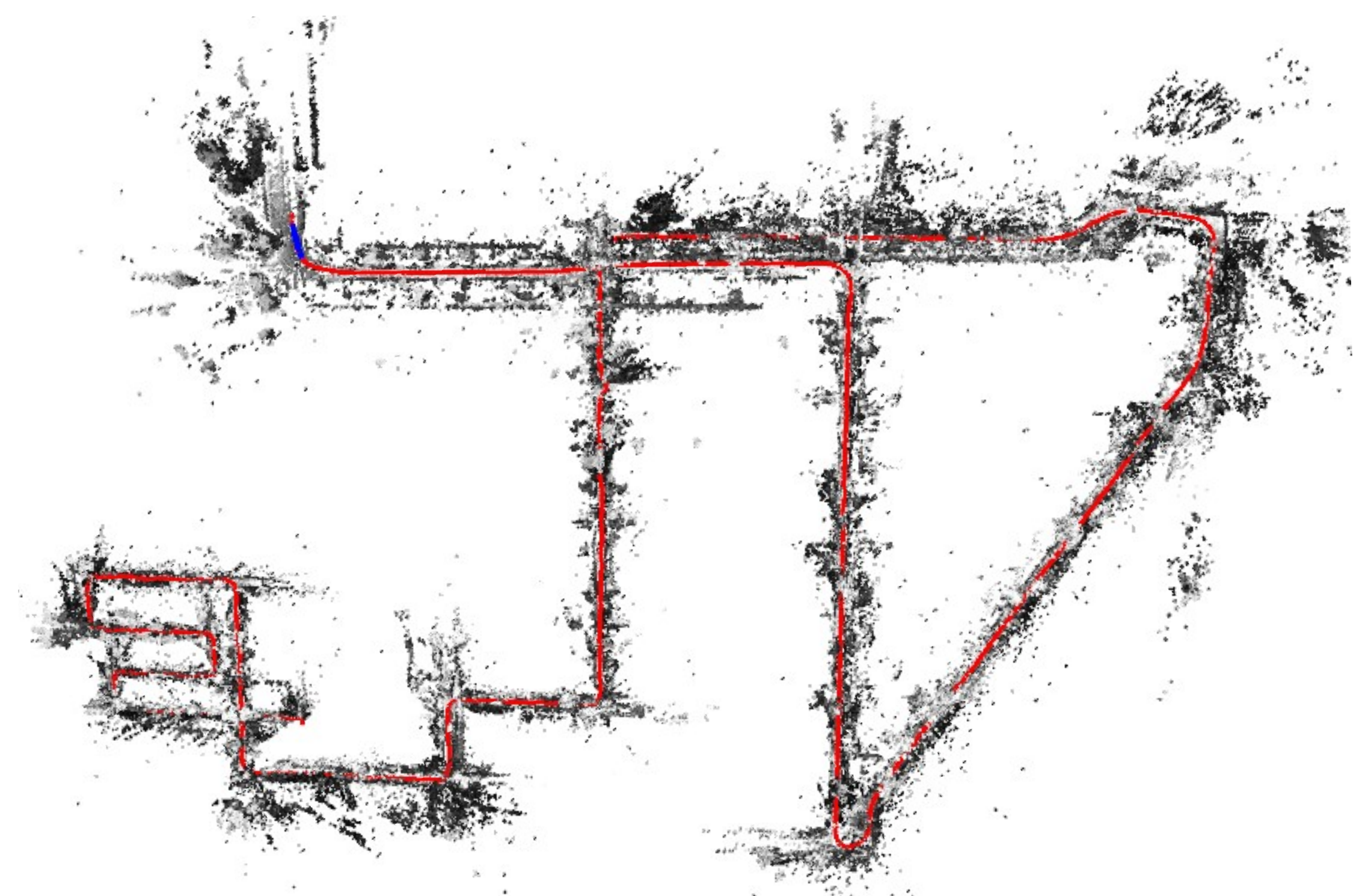}
	}
	\subfigure[Trajectory generated by DDSO on Seq. 09]{
		\includegraphics[width = 0.47\columnwidth]{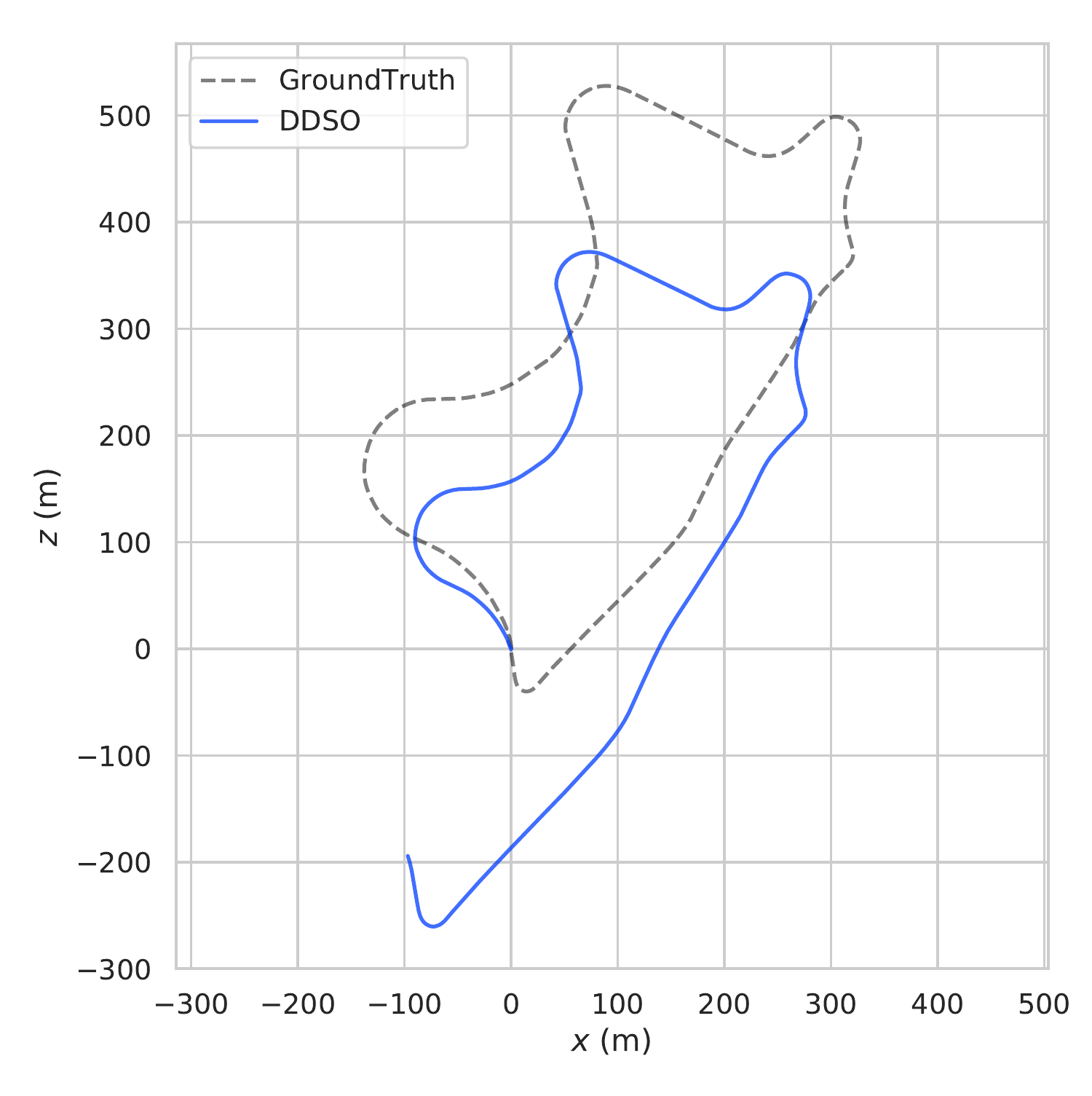}
	}
	\subfigure[The point cloud on Seq. 09]{
		\includegraphics[width = 0.47\columnwidth]{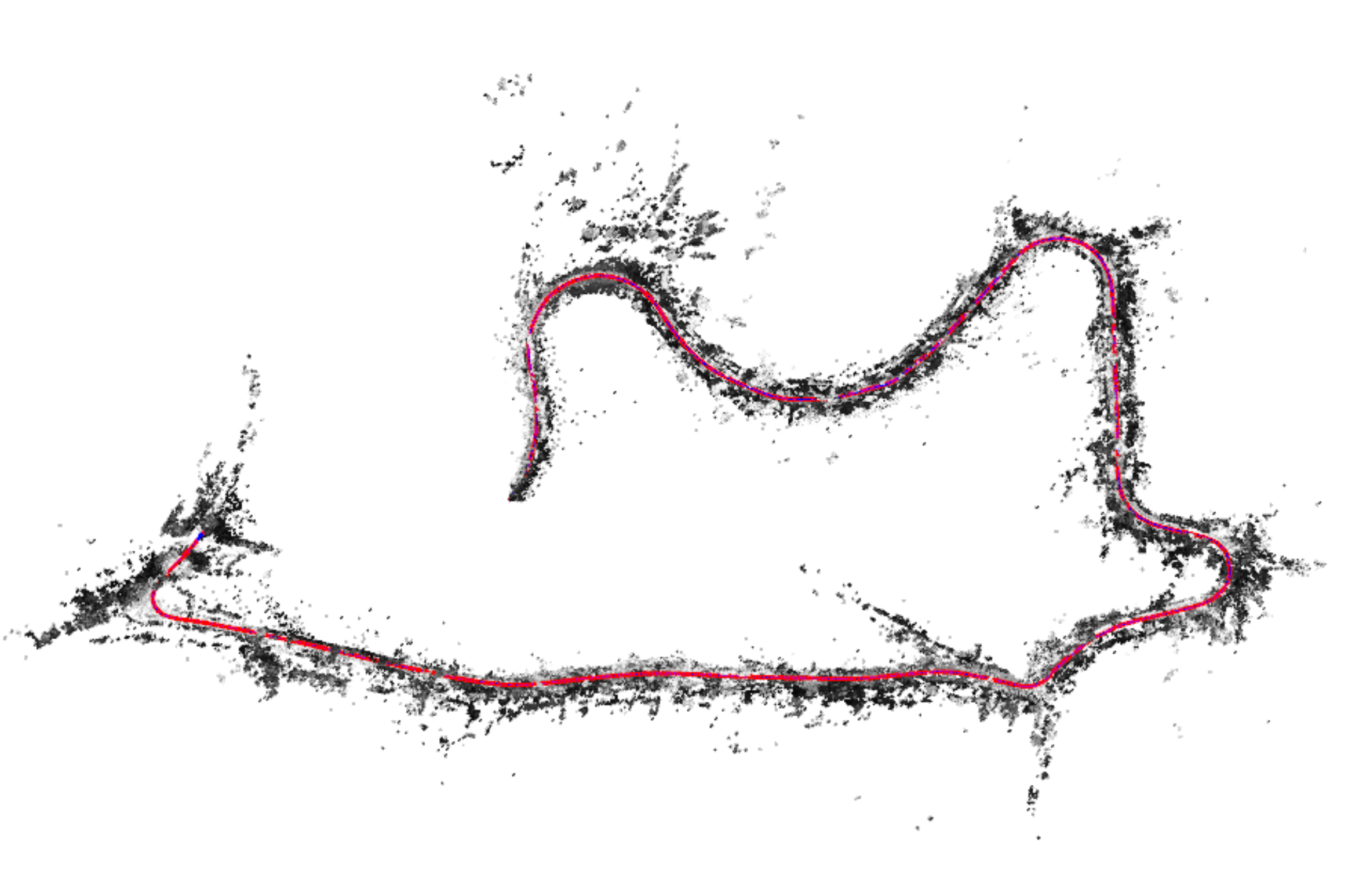}
	}
	\subfigure[Trajectory generated by DDSO on Seq. 10]{
		\includegraphics[width = 0.47\columnwidth]{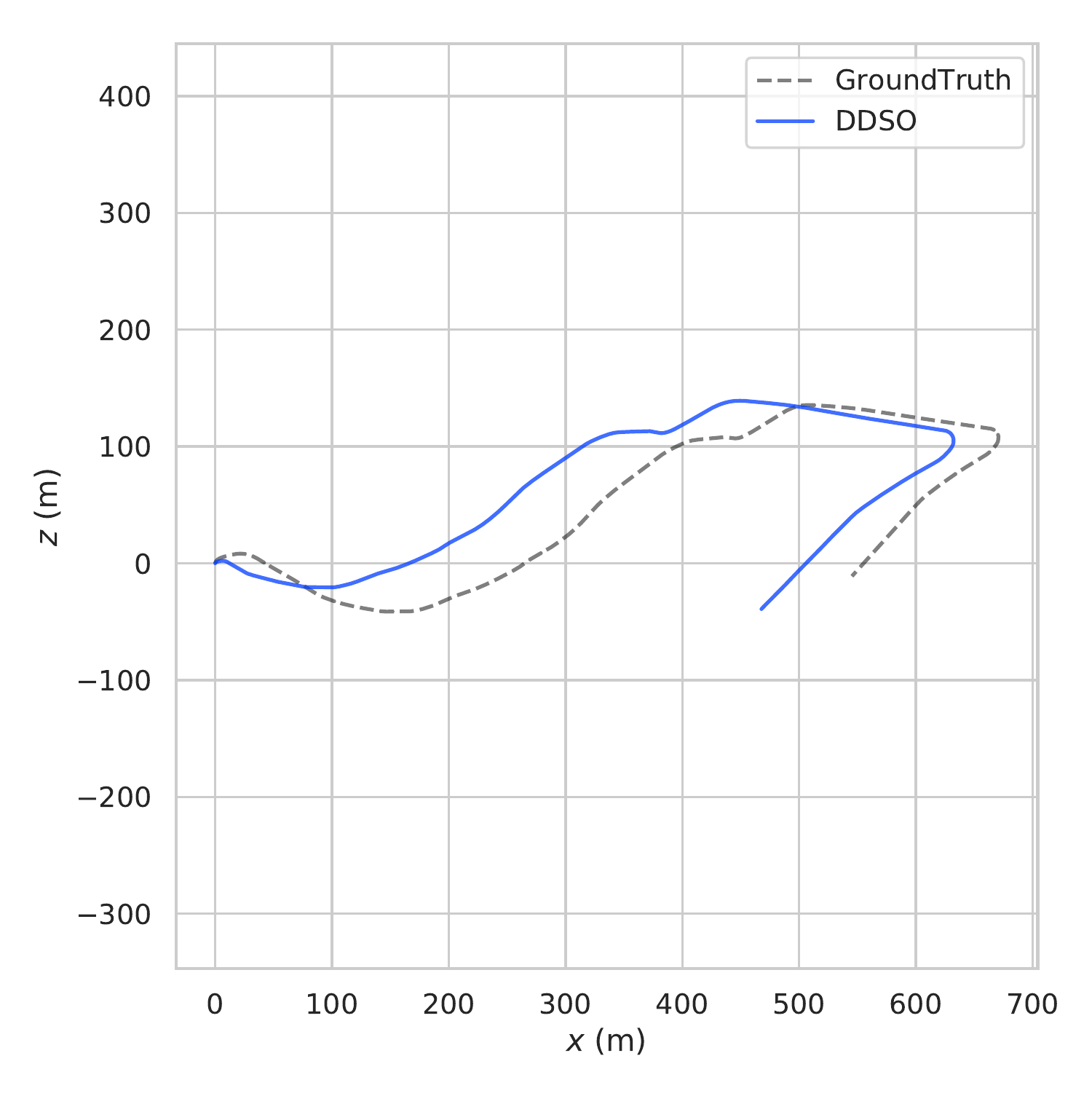}
	}
	\subfigure[The point cloud on Seq. 10]{
		\includegraphics[width = 0.47\columnwidth]{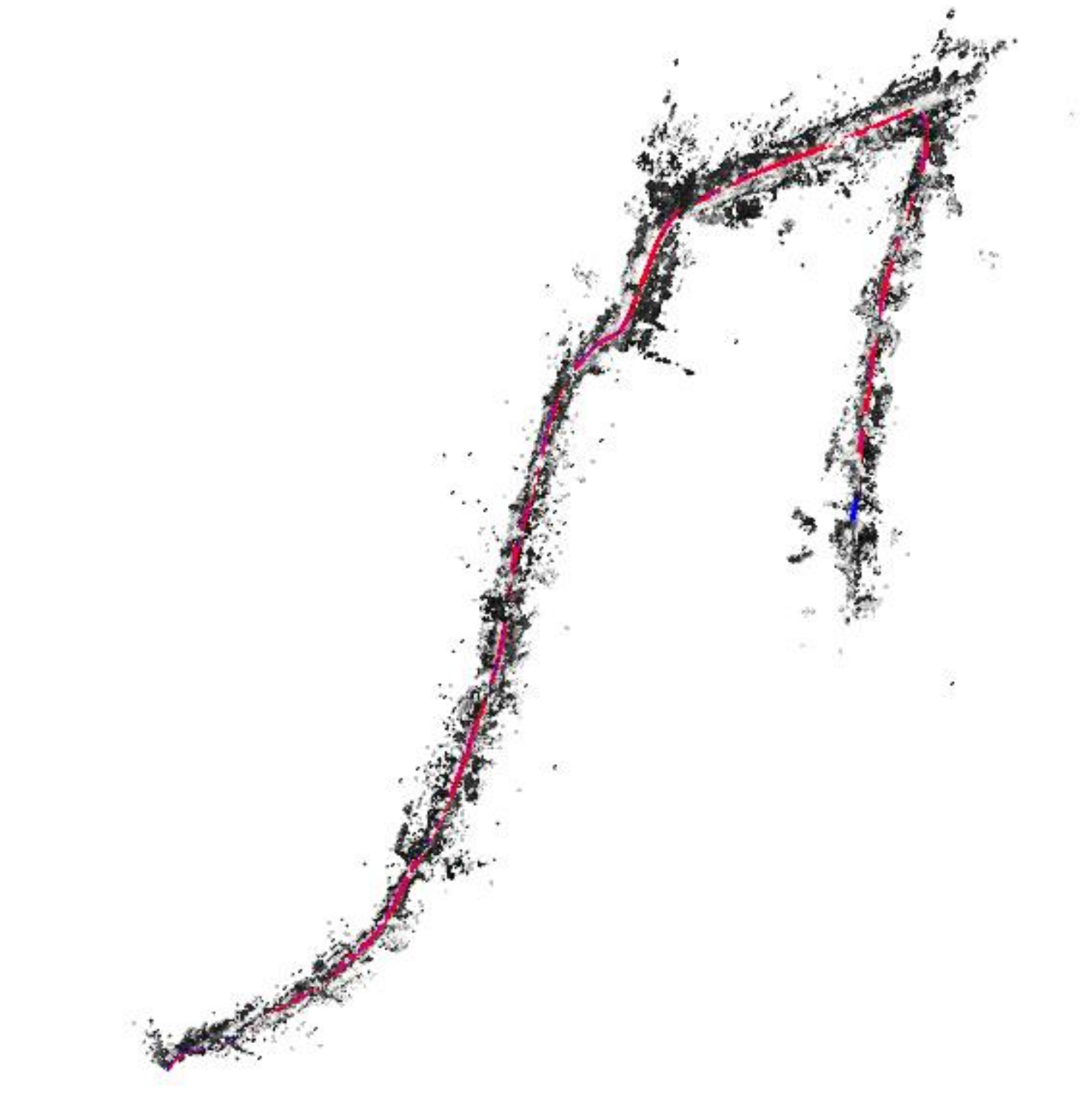}
	}
	\caption{ Trajectories and their point clouds generated by DDSO on KITTI odometry sequences 07-10. Although DDSO suffers from the widespread scale-drift in monocular VO, DDSO has the mapping thread to reconstruct the 3D map of the environment when compared with deep learning-based VO.}
	\label{fig:fig7}
\end{figure*}

\begin{figure*}[htbp]
	\centering
	\subfigure[Trajectories on Seq. 07]{
		\includegraphics[width = 0.47\columnwidth]{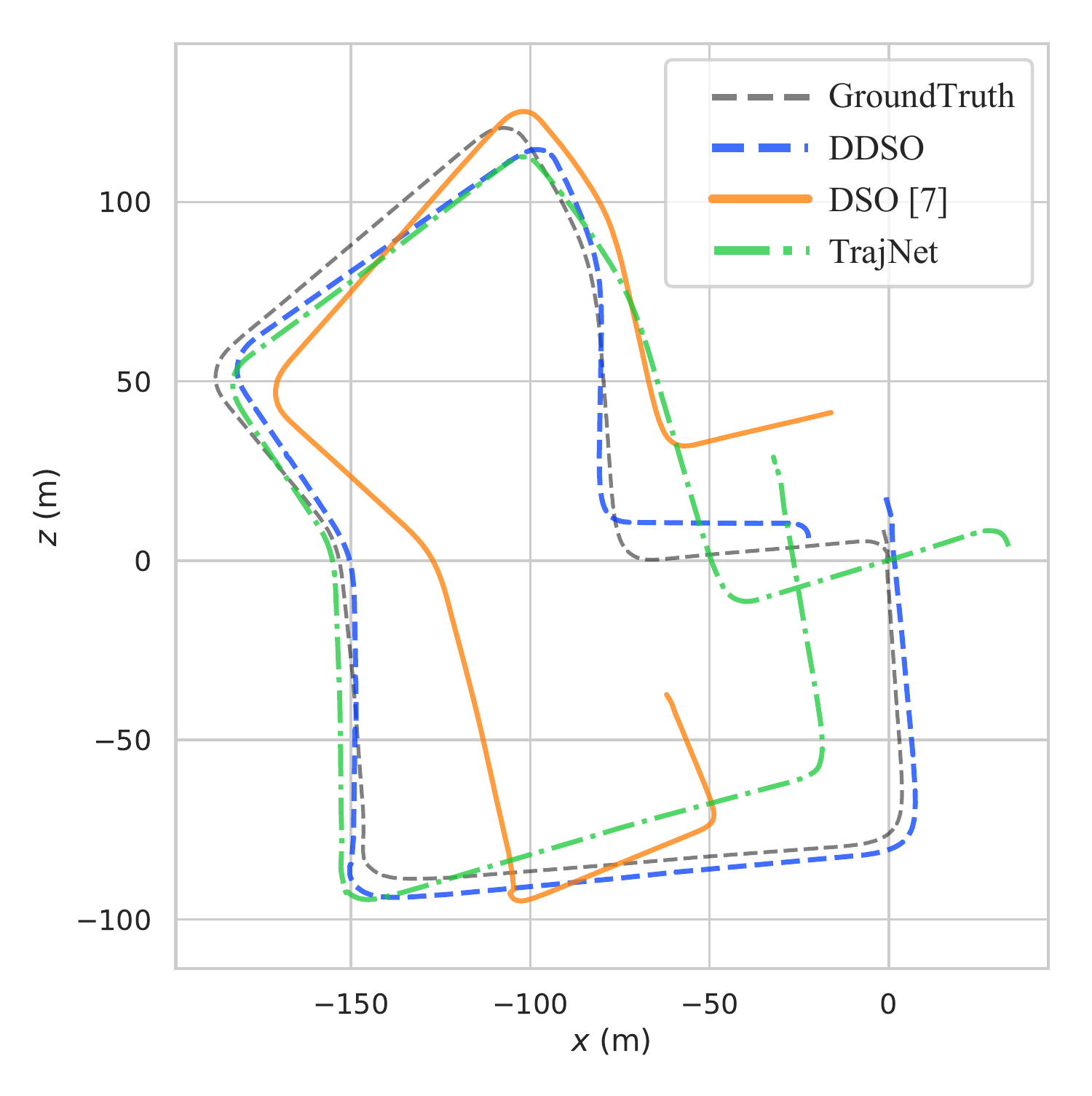}
	}
	\subfigure[Trajectories on Seq. 08]{
		\includegraphics[width = 0.47\columnwidth]{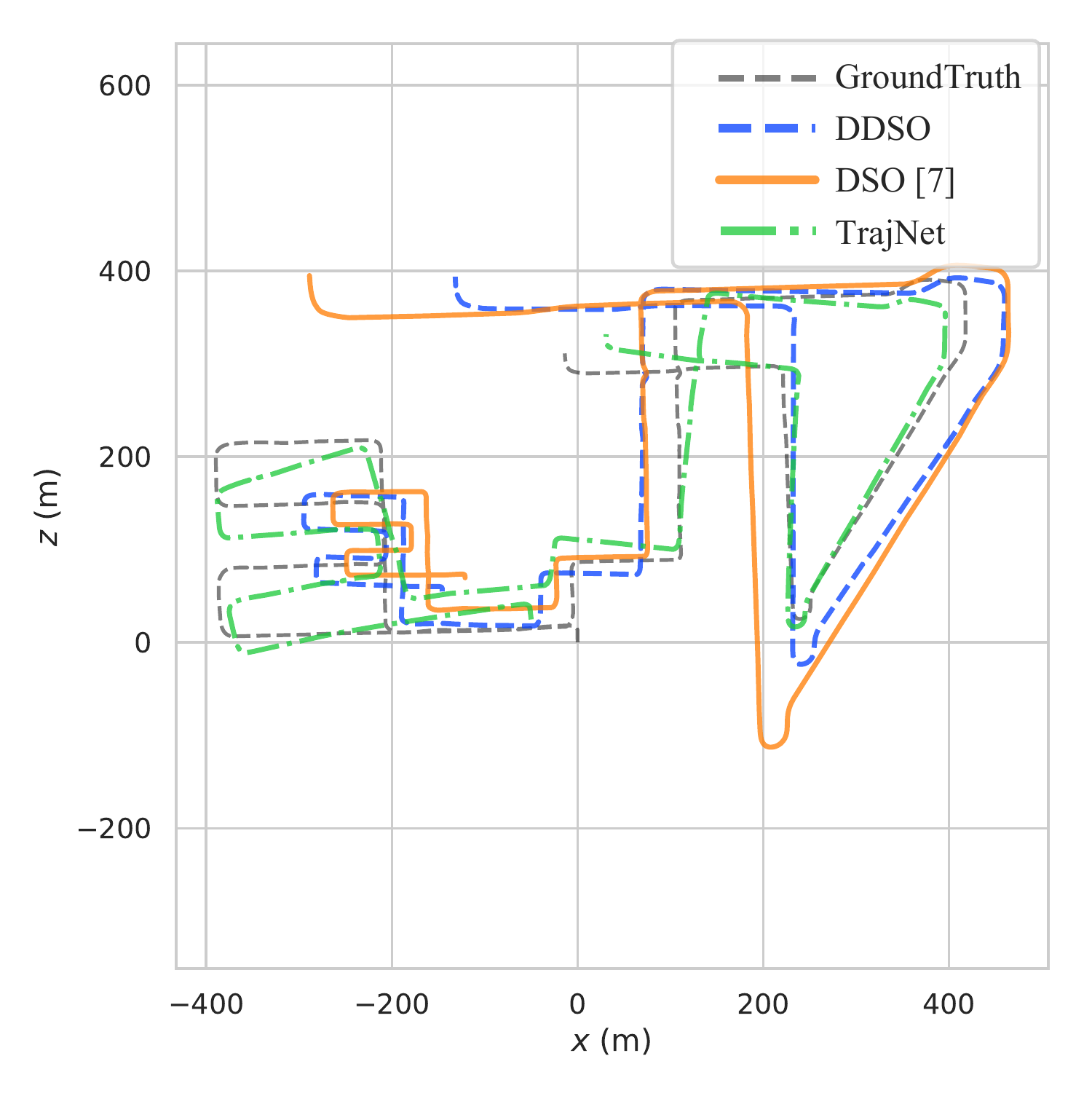}
	}
	\subfigure[Trajectories on Seq. 09]{
		\includegraphics[width = 0.47\columnwidth]{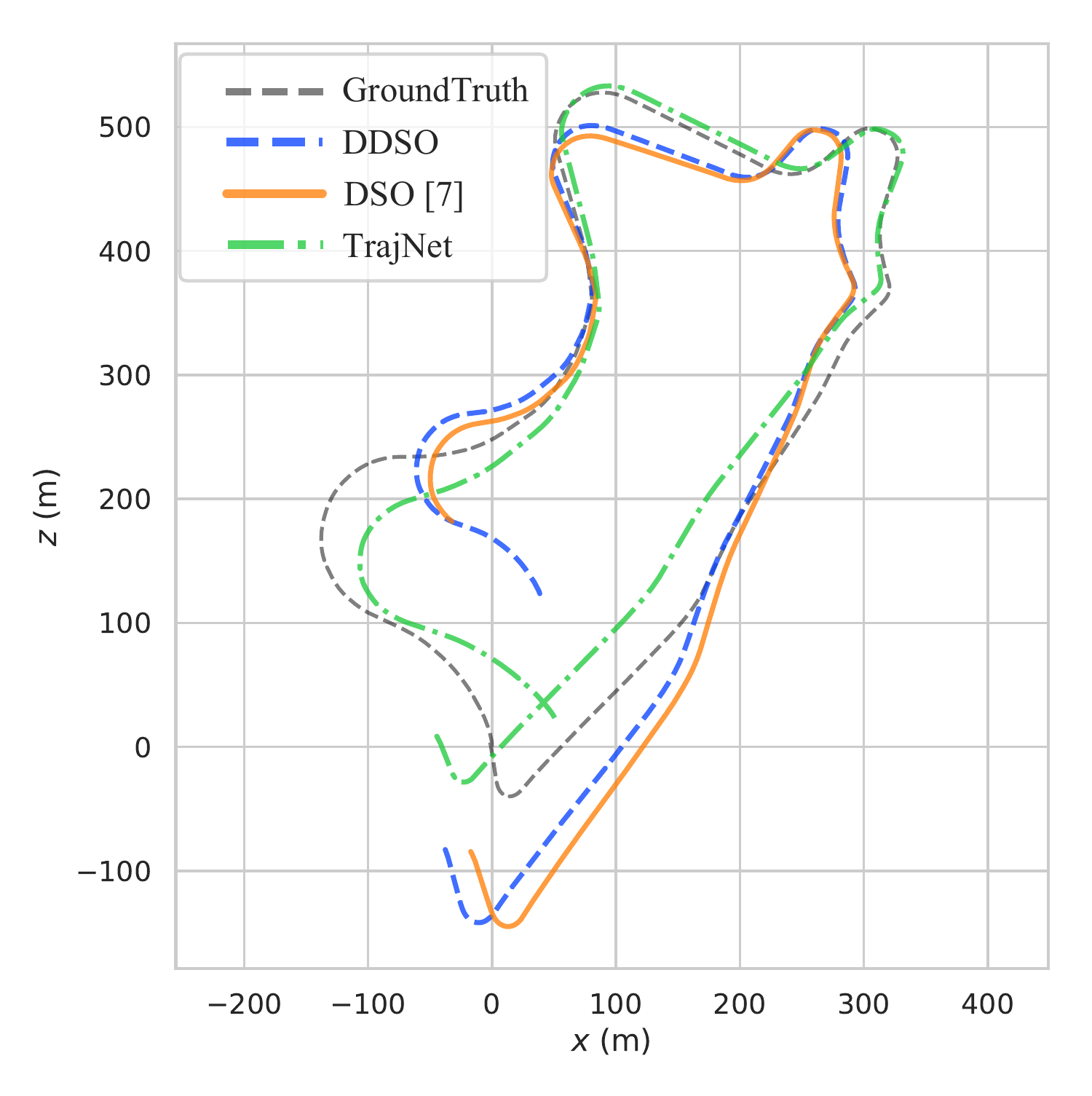}
	}
	\subfigure[Trajectories on Seq. 10]{
		\includegraphics[width = 0.47\columnwidth]{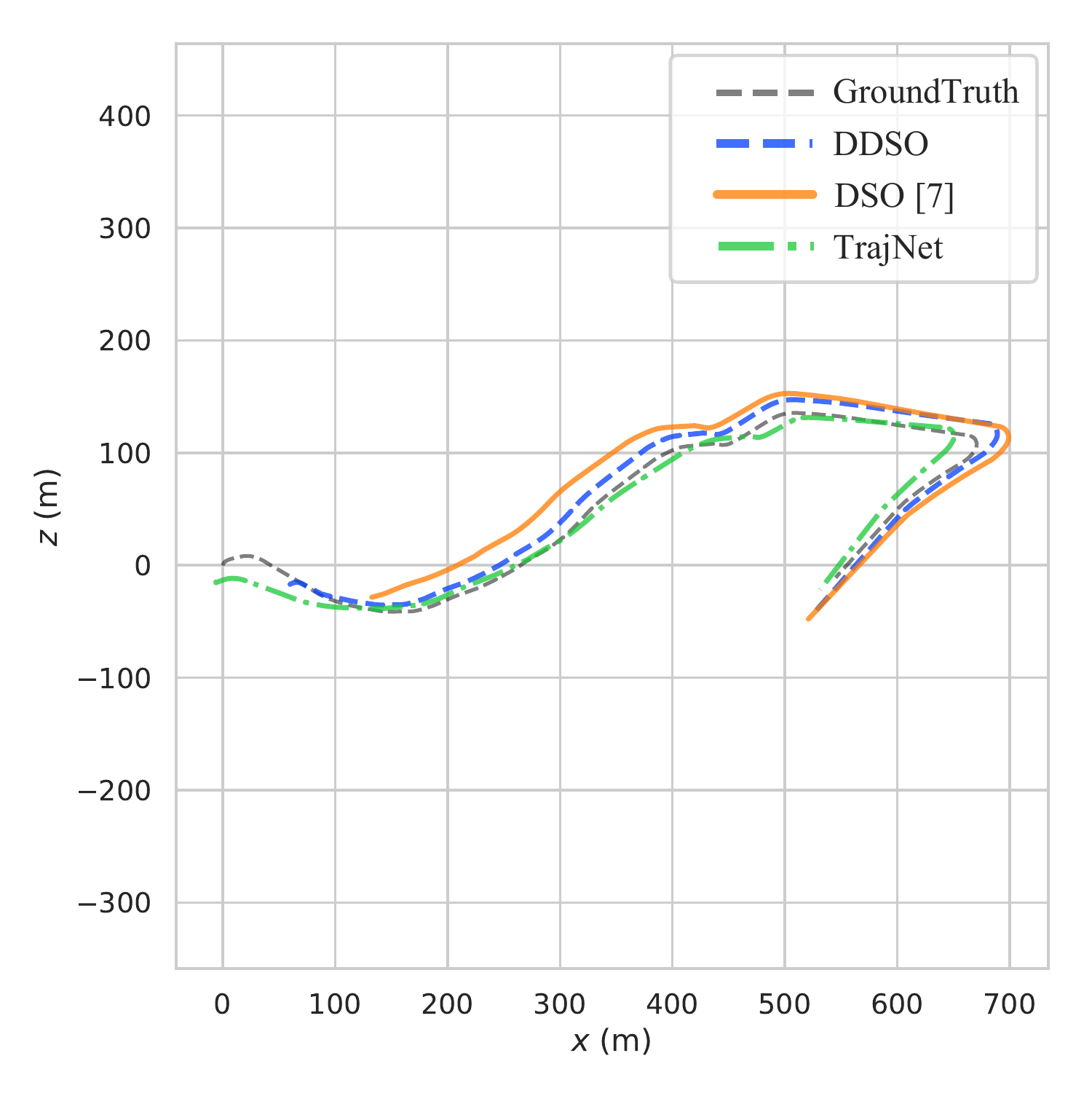}
	}
	\caption{ Sample trajectories comparing DDSO, DSO \cite{engel2017direct} and TrajNet, and the ground truth on a metric scale. DDSO shows a better odometry estimation in terms of both initialization and tracking than DSO \cite{engel2017direct}. In addition, compared with DSO \cite{engel2017direct} and DDSO, the trajectories generated by TrajNet do not suffer from the scale-drift problem, which is the main source of error in traditional monocular VO.}
	\label{fig:fig8}
\end{figure*}

\textbf{Comparison of different methods: }We evaluate the performance of TrajNet on KITTI VO sequences 09-10, and the results are shown in Fig. \ref{fig:fig5}. The standard evaluation tools provided by the KITTI dataset are used to evaluate the full predicted trajectories, which is different from the 5-frame-trajectory evaluation used in \cite{zhou2017unsupervised,yin2018geonet}. Table \ref{Tab02} shows the average rotation and translation errors of predicted trajectories on the KITTI odometry sequences 09 and 10. Note that since monocular depth estimation methods based on stereo image sequences \cite{zhan2018unsupervised} can learn the scale information from stereo image pairs, their pose network does not suffer from scale-inconsistency and scale-ambiguity problems. Because of the scale ambiguity, the scale information of trajectories predicted by monocular methods \cite{zhou2017unsupervised,yin2018geonet,zhan2018unsupervised,bian2019depth,wang2019recurrent,zhao2020masked} are autonomously aligned with reference to the ground truth before evaluation, as shown in Fig. \ref{fig:fig6}. Compared with previous methods in full trajectory prediction, the proposed method shows outstanding performance in both translation and orientation prediction over a long monocular video, and the qualitative results are shown in Fig. \ref{fig:fig6}.

\subsection{Deep Direct Sparse Odometry}

For DDSO, we compare its initialization process as well as tracking accuracy on the odometry sequences of the KITTI dataset \cite{geiger2013vision} against the original DSO \cite{engel2017direct} (without photometric camera calibration), as shown in Fig. \ref{fig:fig1} and Fig. \ref{fig:fig8}. The python package, evo \footnote[1]{\url{https://github.com/MichaelGrupp/evo}}, is used to evaluate the trajectory errors of DDSO and DSO \cite{engel2017direct}.

We use the KITTI odometry 00-06 sequences for retraining TrajNet and 07-10 sequences for testing on DSO \cite{engel2017direct} and DDSO. Fig. \ref{fig:fig7} shows the point-clouds on sequences 07-10 generated by DDSO and the comparison between DDSO and DSO. This verifies that the proposed DDSO framework works well, and the strategy of replacing pose initialization models including a constant motion model with TrajNet is effective. Then, both the absolute pose error (APE) and relative pose error (RPE) of trajectories generated by DDSO and DSO are computed by the trajectory evaluation tools in evo. As shown in Table \ref{Tab03}, DDSO achieves better performance than DSO on sequences 07-10. 
Table \ref{Tab03} also shows the advantage of DDSO in improving the initialization of DSO. Because of its inability to handle several brightness changes between frames, DSO cannot complete the initialization smoothly and quickly on sequences 07, 09 and 10. However, the photometric has little effect on the pose estimation based on deep learning, and the nonsensical initialization is replaced by the relatively accurate pose estimation regressed by TrajNet during initialization, so that DDSO can finish the initialization successfully and robustly. From the errors of sequence 08 in Table \ref{Tab03}, both DSO and DDSO can successfully achieve initialization, and DDSO shows more accurate tracking than DSO, which confirms the effectiveness of the proposed DDSO framework in improving the accuracy of tracking. Because the widely used constant motion model is replaced by TrajNet during tracking, a more accurate initial pose is provided to the image alignment algorithm, so that DDSO outperforms DSO during tracking.

\begin{table*}[t]
	
	\scriptsize
	
	\centering
	
	\caption{ \textit{Pose Error (in metric) on KITTI Dataset \cite{geiger2013vision}}}
	
	\label{Tab03}
	
	\begin{tabular}{cc|ccccc|ccccc|c}
		
		\toprule
		
		\multirow{2}{*}{Sequence}& \multicolumn{6}{c}{$Absolute$ $Pose$ $Error^{2}$ $(APE)$} & \multicolumn{6}{c}{$Relative$ $Pose$ $Error^{2}$ $(RPE)$}  \\
		
		\cmidrule(r){2-7} \cmidrule(r){8-13}
		
		&  $Method$   &   $Max$ 	&  $Mean$     &   $Min$		&  $Rmse$      &  $Std$    &   $Max$ 	&  $Mean$     &   $Min$		&  $Rmse$      &  $Std$     &  $ICD^{1}$      \\
		
		\cline{1-13}

		\multirow{2}{*}{seq 07}
		
		&  $DSO$ \cite{engel2017direct}      &  76.35  &   33.10 &   \textbf{1.19}  &   38.22   &    19.10   &  2.68  &   0.20 	 &   \textbf{0.001}  &   0.287   &    0.205  &  $False$    \\
		
		
		&  $DDSO$      &  \textbf{23.28}   &   \textbf{7.92}  &   1.41 &   \textbf{9.06}	&    \textbf{4.40}  &  \textbf{0.68}  &   \textbf{0.07}  &   \textbf{0.001}	&    \textbf{0.096} 	&   \textbf{0.063 } &  $\textbf{True}$    \\
		
		\midrule
		
		\multirow{2}{*}{seq 08}
		
		&  $DSO$  \cite{engel2017direct}     &  287.44  &   74.89 &   10.62  &   92.84   &    54.865   &  1.44  &   0.386 	 &  0.002  &   0.473   &    0.273 &  $\textbf{True}$    \\
		
		
		&  $DDSO$      &  \textbf{155.85}   &   \textbf{53.96}  &   \textbf{1.62 } &   \textbf{63.91}	&    \textbf{34.26}  &  \textbf{0.92}  &   \textbf{0.299} 	&   \textbf{0.001 } &   \textbf{0.367}	&    \textbf{0.213}  &  $\textbf{True}$    \\
		
		\midrule
		\midrule
		
		\multirow{2}{*}{seq 09}
		
		&  $DSO$ \cite{engel2017direct}      &  184.96  &   57.57  &   2.83  &   69.16   &    38.33   &  3.168  &   0.302 	 &   \textbf{0.002}  &   0.418   &    0.288  &  $False$    \\
		
		
		&  $DDSO$      &  \textbf{129.98}   &   \textbf{53.40}  &   \textbf{1.59 } &   \textbf{62.72}	&    \textbf{32.90}  &  \textbf{0.839}  &   \textbf{0.229} 	&   \textbf{0.002 } &   \textbf{0.254}	&    \textbf{0.111}  &  $\textbf{True}$    \\
		
		\midrule
		
		\multirow{2}{*}{seq 10}
		
		&  $DSO$ \cite{engel2017direct}      &  135.66  &   35.53  &   10.86  &   44.12   &    26.14  &  4.006  &   0.286 	 &   0.003  &   0.421   &    0.309  &  $False$    \\
		
		
		&  $DDSO$      &  \textbf{62.24} 	&   \textbf{16.46}    &   \textbf{0.96}  &   \textbf{20.31}	&    \textbf{11.89 }  &  \textbf{0.734}  &   \textbf{0.139} 	&   \textbf{0.001}  &   \textbf{0.227}	&    \textbf{0.179}    &  $\textbf{True}$    \\
		
		\bottomrule
		
	\end{tabular}
	
	\footnotesize{$^{1}$ ICD means whether the initialization can be completed within the first 20 frames}\\
	
	\footnotesize{$^{2}$ Lower is better}\\
	
\end{table*}

\begin{table*}[]
	
	\scriptsize
	
	\centering
	
	\caption{ \textit{Comparison between DSO \cite{engel2017direct}, DDSO and TrajNet} }
	
	\label{Tab04}
	\resizebox{1.6\columnwidth}{!}{
		\begin{tabular}{c|cc|cc|cc|cc}
			
			\toprule
			\multicolumn{1}{c}{}& \multicolumn{2}{c}{Seq. 07} & \multicolumn{2}{c}{Seq. 08}& \multicolumn{2}{c}{Seq. 09} & \multicolumn{2}{c}{Seq. 10}  \\
			\cmidrule(r){2-3} \cmidrule(r){4-5}\cmidrule(r){6-7} \cmidrule(r){8-9}
			
			Method 	& $t_{err}(\%)$ & $r_{err}(^{\circ}/100m)$ & $t_{err}(\%)$ & $r_{err}(^{\circ}/100m)$ & $t_{err}(\%)$ & $r_{err}(^{\circ}/100m)$ & $t_{err}(\%)$ & $r_{err}(^{\circ}/100m)$  \\
            \hline
            $TrajNet^{1}$  &  10.93  &   5.15  &   \textbf{8.73}  &   2.32   &    \textbf{8.35}  &  3.00  &   11.90 	 &   2.87 \\
	
            $DSO$ \cite{engel2017direct}     & 22.33 & 3.98 & 24.85 & \textbf{0.25} & 18.74 & 1.29 & 28.16 & 4.68 \\
            $DDSO$     &  \textbf{6.56}  &   \textbf{1.23}  &   19.37  &   \textbf{0.25}   &    15.02  &  \textbf{0.20}  &   \textbf{8.76} 	 &   \textbf{0.66} \\
			
			\hline

			\bottomrule
			
		\end{tabular}
	}
\\
	   \footnotesize{$^{1}$ The TrajNet here is trained on sequences 00-06, which is different with that in Tables \ref{Tab01} and \ref{Tab02}.}\\
\end{table*}

\subsection{Comparison of DSO \cite{engel2017direct}, DDSO and TrajNet:}
 To argue for the advantages of different deep learning-based frameworks, we compare traditional-based DSO \cite{engel2017direct}, deep learning-based TrajNet and fusion-based DDSO. The trajectories generated by different frameworks on KITTI odometry sequences 07-10 are shown in Fig. \ref{fig:fig8} (a)-(d). Due to its sensitivity to photometry, DSO cannot obtain robust initialization in HDR environments without photometric camera calibration. If traditional VO cannot successfully finish its initialization, the follow-up tracking process will also be influenced. The deep learning-based TrajNet proposed in this paper shows an outstanding ability in full trajectory generation over long monocular videos (e.g., sequence 08 has more than 4,000 frames). Compared with DSO and DDSO, TrajNet hardly suffers from scale-drift, which is a widespread problem and the main source of error in monocular VOs. However, as shown in Fig. \ref{fig:fig8} and Table \ref{Tab04}, TrajNet is not accurate enough in orientation estimation, especially the large direction change between frames. In addition, considering the lack of mapping threads, the deep learning-based VO methods cannot provide the 3D geometric prior information of environments for autonomous systems, so their applications will be limited. The integration of TrajNet and DSO perfectly solves some of the above problems, such as the robust initialization and accurate tracking of DSO, and the mapping of TrajNet.
 As shown in Fig. \ref{fig:fig8}, with the help of TrajNet, DDSO achieves robust initialization and more accurate tracking than DSO. Moreover, the cooperation with traditional methods also provides a direction for the practical application of the current learning-based pose estimation. As shown in Table \ref{Tab04}, DDSO has better performance in orientation estimation than TrajNet. Moreover, the 3D map (point cloud) of surroundings can be reconstructed by DDSO, and it is helpful for the environmental perception and decision-making of robots, such as the subsequent navigation and obstacle avoidance. However, the scale-drift problem in DDSO is not effectively constrained.

\section{CONCLUSION}

In this paper, we study the application of deep learning in the ego-motion prediction of robots and vehicles, including inferring the full trajectory of monocular video by TrajNet and improving traditional DSO by combining TrajNet.
We address the scale-inconsistency problem that widely exists in previous pose networks by providing novel constraints, and TrajNet is trained by monocular videos in an unsupervised manner, which is free from ground truth labels. Experiments show the effectiveness of the proposed novel constraint, and TrajNet outperforms previous works in full trajectory prediction.
In addition, we present a novel monocular DVO framework, DDSO, which incorporates TrajNet into the traditional DSO framework \cite{engel2017direct}. The initialization and tracking are improved by using the TrajNet output as an initial value in the image alignment algorithm. The evaluation conducted on the KITTI odometry dataset demonstrates that DDSO outperforms the original monocular DSO \cite{engel2017direct}. Meanwhile, the proposed DDSO achieves more robust initialization and accurate tracking than DSO \cite{engel2017direct}. The key benefit of the proposed DDSO framework is that it allows us to obtain robust and accurate DSO without complex photometric calibration. Moreover, this idea can also be used in other direct or semi-direct methods which solve poses by image alignment algorithms. Nevertheless, there are still shortcomings that need to be addressed in the future. For example, scale drift still exists in the proposed DDSO, and we plan to train TrajNet on stereo image sequences to help TrajNet learn the absolute scale information, thereby improving the scale ambiguity and scale drift of DDSO.


\ifCLASSOPTIONcaptionsoff
  \newpage
\fi

{
	\bibliographystyle{IEEEtran}
	\bibliography{egbib}
}


\end{document}